\documentclass{article}
\PassOptionsToPackage{square,comma,numbers,compress}{natbib}

\usepackage[preprint]{neurips_2023}

\usepackage[utf8]{inputenc} %
\usepackage[T1]{fontenc}    %
\usepackage[hidelinks,colorlinks=true]{hyperref}       %
\usepackage[dvipsnames]{xcolor}
\usepackage{url}            %
\usepackage{booktabs}       %
\usepackage{amsfonts}       %
\usepackage{nicefrac}       %
\usepackage{microtype}      %
\usepackage{xcolor}         %
\usepackage{graphicx}
\graphicspath{ {figures/} }
\usepackage{array}
\usepackage{times,graphicx,tabulary,multirow,xspace}
\usepackage{subfig}
\usepackage{changepage}
\usepackage{color, colortbl}

\title{GPT-4V Explorations: Mining Autonomous Driving}

\author{
{\bf Zixuan Li$^{1}$} \\ \\
$^{1}$~Waytous, Beijing, China \\ \\
\\
}

\begin{document}

\maketitle

% \vspace{10pt}
\begin{abstract}
This paper explores the application of the GPT-4V(ision) large visual language model to autonomous driving in mining environments, where traditional systems often falter in understanding intentions and making accurate decisions during emergencies. GPT-4V introduces capabilities for visual question answering and complex scene comprehension, addressing challenges in these specialized settings. Our evaluation focuses on its proficiency in scene understanding, reasoning, and driving functions, with specific tests on its ability to recognize and interpret elements such as pedestrians, various vehicles, and traffic devices. While GPT-4V showed robust comprehension and decision-making skills, it faced difficulties in accurately identifying specific vehicle types and managing dynamic interactions. Despite these challenges, its effective navigation and strategic decision-making demonstrate its potential as a reliable agent for autonomous driving in the complex conditions of mining environments, highlighting its adaptability and operational viability in industrial settings.
\end{abstract}

%{\bf keywords: GPT-4V, Scenario understanding, Reasoning, Act as a driver}
\begin{figure*}[h]
\centering
\includegraphics[width=\linewidth]{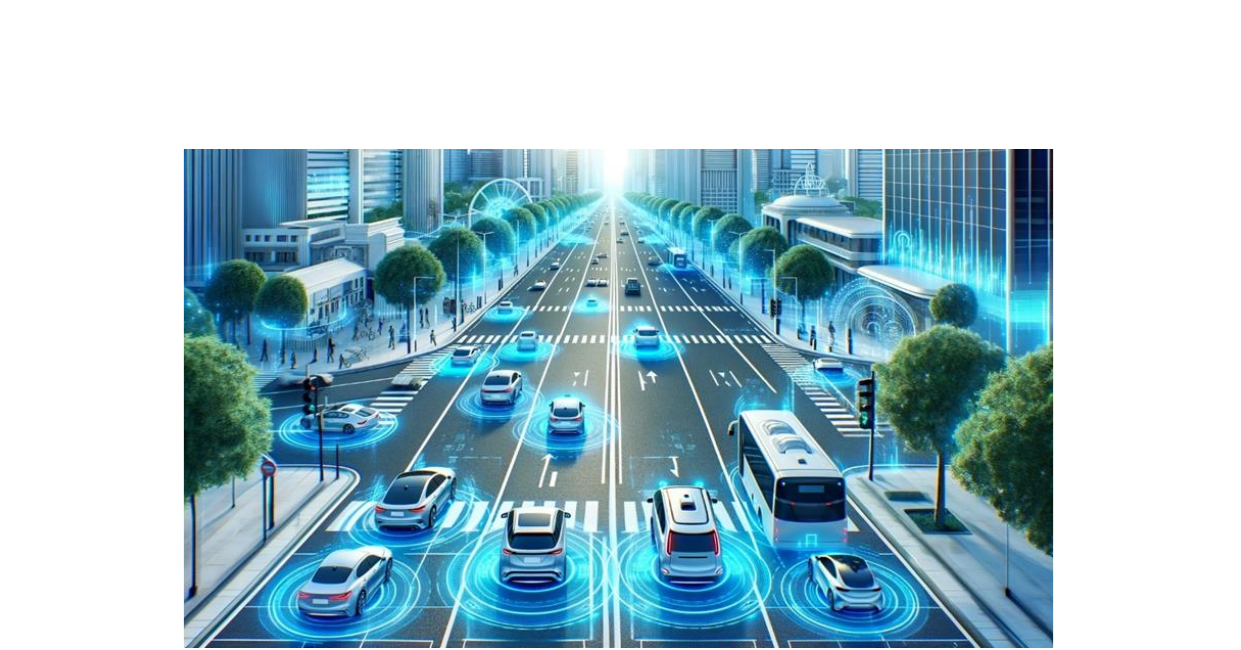}
\caption{An illustration showing the integration of visual language models such as GPT-4V. This picture is generated by DALL·E 3.}
\label{fig_0}
\end{figure*}

\clearpage
{
  \hypersetup{linkcolor=black}
  \tableofcontents
  \label{sec:toc}
}

\clearpage
\section{Introduction}
Autonomous driving technologies are developing at a rapid pace. Many algorithms about autonomous driving, when tested on public datasets, have achieved performance levels that satisfy the criteria for Level 3 and even Level 4 autonomy. Despite these advancements, traditional autonomous driving systems continue to depend on a structured approach encompassing perception, control, and decision-making processes to facilitate vehicle navigation.Nonetheless, current autonomous driving algorithms encounter significant challenges. These algorithms often exhibit long-tail effects and limited generalization due to the restricted diversity of object types in training datasets. Moreover, they typically lack a robust understanding of the intentions of the surrounding entities, particularly in specialized contexts like mining scenarios. This deficiency becomes particularly evident in emergency or extreme conditions, where the algorithms struggle to perform accurate reasoning and to make swift, appropriate decisions\cite{li2024miningllm,teng2024scenario,li2024unstrprompt}.

Incorporating large language models like Llama2\cite{touvron2023llama}, Gemini\cite{gemini,lee2023gemini}, and GLM\cite{du2022glm} into autonomous driving systems represents a promising strategy to address existing challenges\cite{fu2023challenger,r1}. Although these models demonstrate considerable environmental comprehension, their effectiveness in specific contexts such as mining scenarios, where off-road vehicle state inference and path planning are essential, still requires thorough evaluation. Effective reasoning and decision-making, facilitated by these models, are crucial to ensuring the stable operation of autonomous vehicles in such complex environments.

GPT-4V\cite{gpt4,gpt4v_2,gpt4v_3}, a pioneering visual large language model, introduces an innovative approach to autonomous driving in mining settings. This model is equipped with capabilities for both visual Question answering and complex scene comprehension. The focus of this research is to conduct an experimental assessment of GPT-4V's scene understanding and decision-making abilities in the specific application of mining autonomous driving\cite{fu2023challenger,guo2024remote,wu2023gpt4vis}. A battery of tests, meticulously designed to evaluate the model's adaptability and effectiveness in demanding mining environments, was implemented. This paper provides a detailed exploration of GPT-4V’s potential and limitations in autonomous driving, presenting systematic analyses of its performance in varied scenarios\cite{wu2023early,shi2023exploring}.

Our evaluation of GPT-4V's capabilities is structured around three critical dimensions: scene understanding, reasoning, and its ability to perform driving functions. These elements are crucial for assessing the model's effectiveness in managing the intricacies of autonomous driving in challenging mining environments. By scrutinizing how GPT-4V handles these aspects, we aim to validate its potential and reliability as an autonomous driving agent, particularly in situations that require sophisticated cognitive and operational skills.

Scenario Understanding: This evaluation is designed to assess CPT-4V's object recognition capabilities within mining environments, targeting a range of elements including pedestrians, vehicles of various types, mechanical facilities, ore piles, and traffic control devices such as signals and road signs. It also examines the attributes of these objects, such as their distance, position, speed, and content. The aim is to gauge CPT-4V's accuracy in identifying and interpreting these diverse elements within the complex backdrop of a mining site. This proficiency is crucial for ensuring safe and effective navigation through such environments. The ability to accurately recognize and respond to environmental cues enables GPT-4V to dynamically interact with its surroundings, thereby adjusting vehicle behaviors based on a comprehensive understanding of the immediate context.

Reasoning: In this section, we assessed GPT-4V’s proficiency in comprehending environmental conditions during emergencies and extreme events, alongside its capability to strategize responses for our vehicle. The evaluation also examined GPT-4V's ability to discern the driving intentions of other vehicles over time. Such assessments are essential to determine GPT-4V's effectiveness in predicting and responding to dynamic and potentially hazardous scenarios, crucial for ensuring the safety and operational efficiency of autonomous vehicles under challenging circumstances.

Act as a driver: In this part of the study, we evaluated GPT-4V's capabilities as a driver within a mining area, assessing its ability to plan routes and execute tasks akin to those performed by skilled human drivers. To this end, we assigned GPT-4V five specific driving tasks: U-turnings, overtaking, pathfinding, parking, and lane merging. For each task, we provided a sequence of images and required GPT-4V to complete the designated maneuvers based on the visual information. This approach enabled us to closely analyze how effectively GPT-4V can interpret complex driving situations and make strategic decisions in real-time, reflecting its potential to function autonomously in diverse and challenging environments.

\section{Scenario Understanding}
\label{sec:scenarioUnderstanding}
There are many complex scenes in mining area, the decision-makers are required to have a comprehensive understanding of the environment and to accurately infer the intentions of other vehicles or facilities present. In our assessment of GPT-4V, specific tests were designed to evaluate its scene understanding capabilities within mining environments. These tests covered a range of functions, including the recognition of various vehicle types such as mining trucks, sedans, and pickups; the identification of road-side signals and signs; and the detection of unstructured roads and ore piles typical of mining areas. The GPT-4V's ability to recognize persons and mining machineries, particularly mechanical arms, was also rigorously tested. For example, when a mechanical arm is in proximity to the vehicle, GPT-4V must closely monitor its movements and respond appropriately, either by slowing down or stopping. Conversely, in more open road scenarios devoid of such obstacles, GPT-4V can employ a more relaxed navigation strategy.

\subsection{Understanding of Persons in Scenes}
\begin{figure*}[h]
    \centering
    \includegraphics[width=\textwidth]{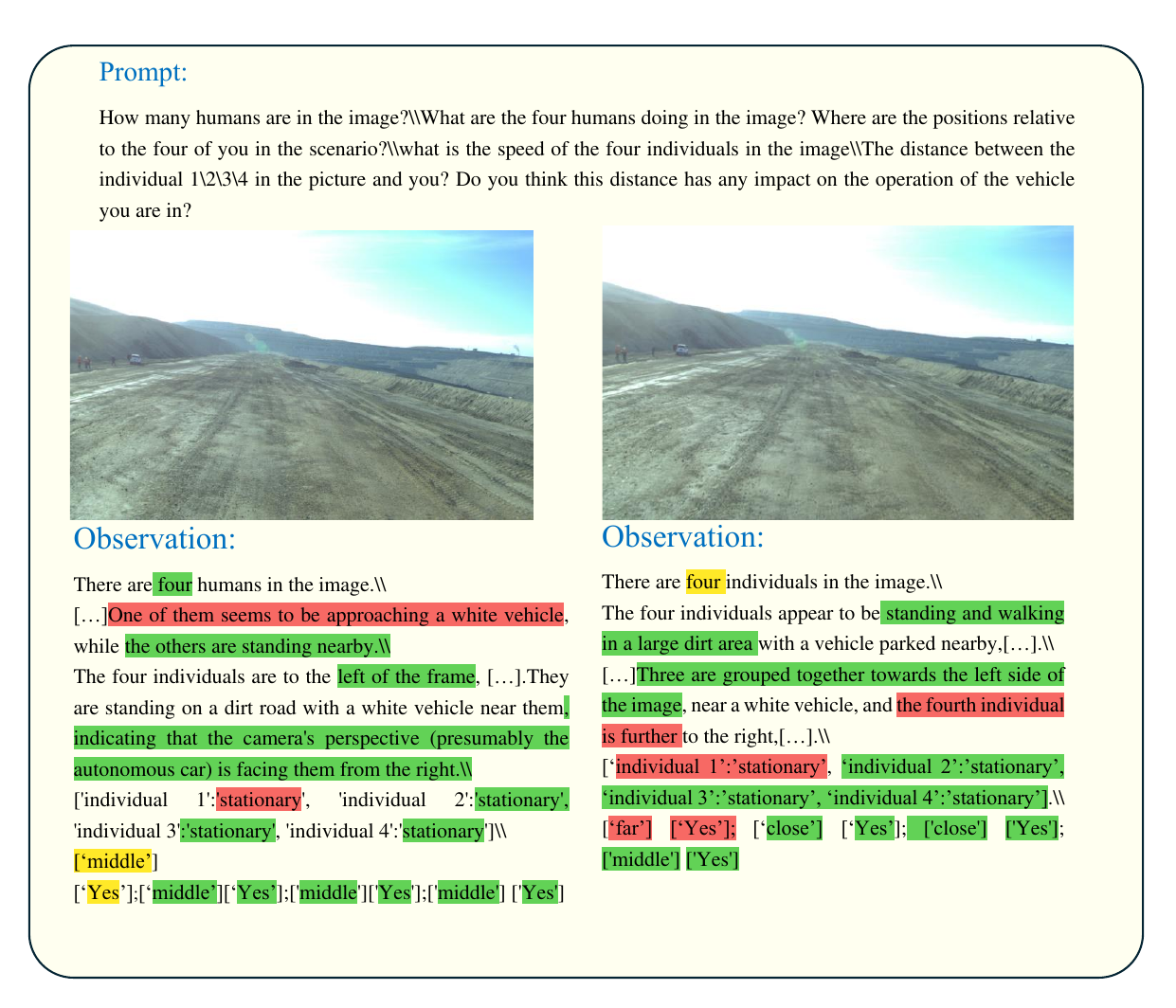}
    \caption{\colorbox{green}{Green}highlights the right answer in understanding,\colorbox{red}{Red}highlights the wrong answer in understanding,\colorbox{yellow}{Yellow} highlights the incompetence in performing the task.}
    \label{fig_1}
\end{figure*}

In evaluating the GPT-4V's understanding ability of persons presence within mining settings, a series of tests were conducted to assess crucial factors including the count of persons, their positioning relative to the vehicle, the proximity between individuals and the vehicle, and their movement states. These elements are crucial given that the dynamics of human movement in mining environments frequently diverge significantly from those observed in urban settings. Although GPT-4V accurately pinpointed individuals' locations from the vehicle, it faced challenges with misidentifications, notably confusing car doors with people in certain cases as illustrated in Figure~\ref{fig_1}. Despite these obstacles, GPT-4V exhibited a generally strong ability to assess human states.
\subsection{Understanding of vehicles in Scenes}
A significant number of mining trucks exited the mine, accompanied by numerous vehicles engaged in construction activities. This scenario presents considerable challenges and is critically important for accurate vehicle driving judgments. In evaluating GPT-4V’s ability in understanding vehicle-related scenes in mining environments, we assessed several parameters: the quantity, type, speed, and direction of vehicles. Overall, GPT-4V exhibited limited accuracy in vehicle recognition within these challenging scenarios.
\begin{figure*}[h]
\centering
\includegraphics[width=\linewidth]{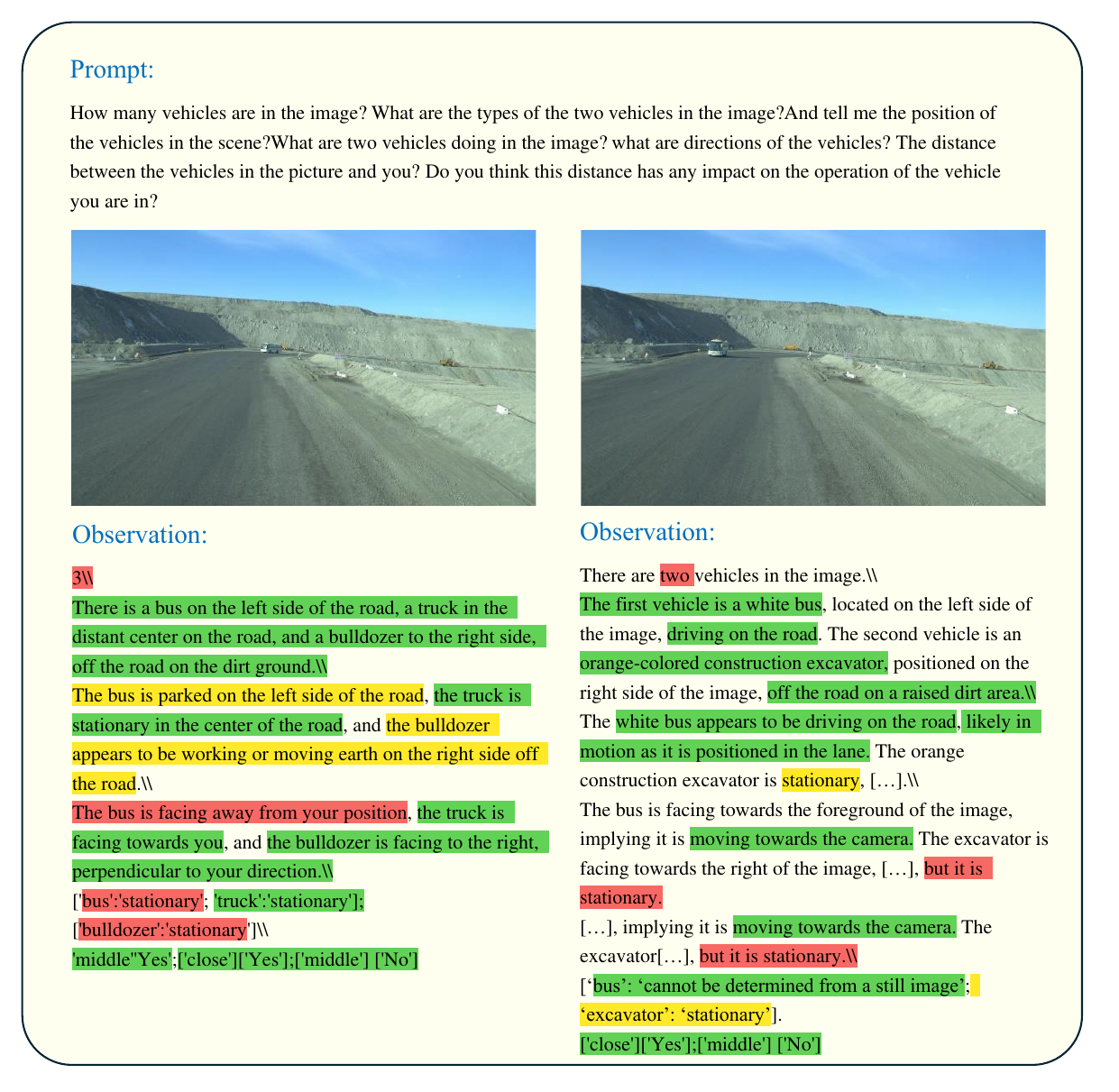}
\caption{\colorbox{green}{Green}highlights the right answer in understanding,\colorbox{red}{Red}highlights the wrong answer in understanding,\colorbox{yellow}{Yellow} highlights the incompetence in performing the task.}
\label{fig_2}
\end{figure*}

\begin{figure*}[h]
\centering
\includegraphics[width=\linewidth]{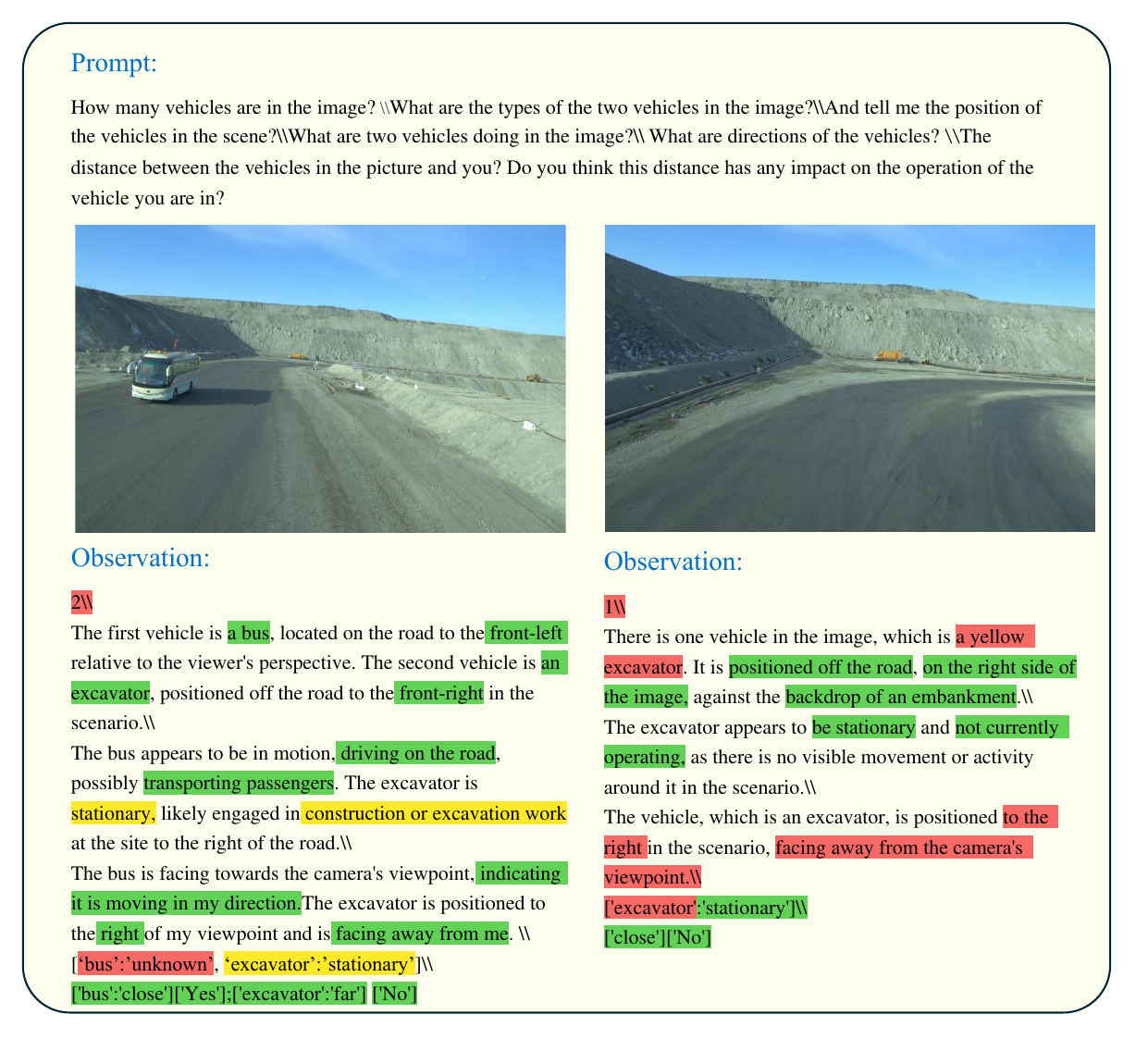}
\caption{\colorbox{green}{Green}highlights the right answer in understanding,\colorbox{red}{Red}highlights the wrong answer in understanding,\colorbox{yellow}{Yellow} highlights the incompetence in performing the task.}
\label{fig_3}
\end{figure*}

\begin{figure*}[h]
\centering
\includegraphics[width=\linewidth]{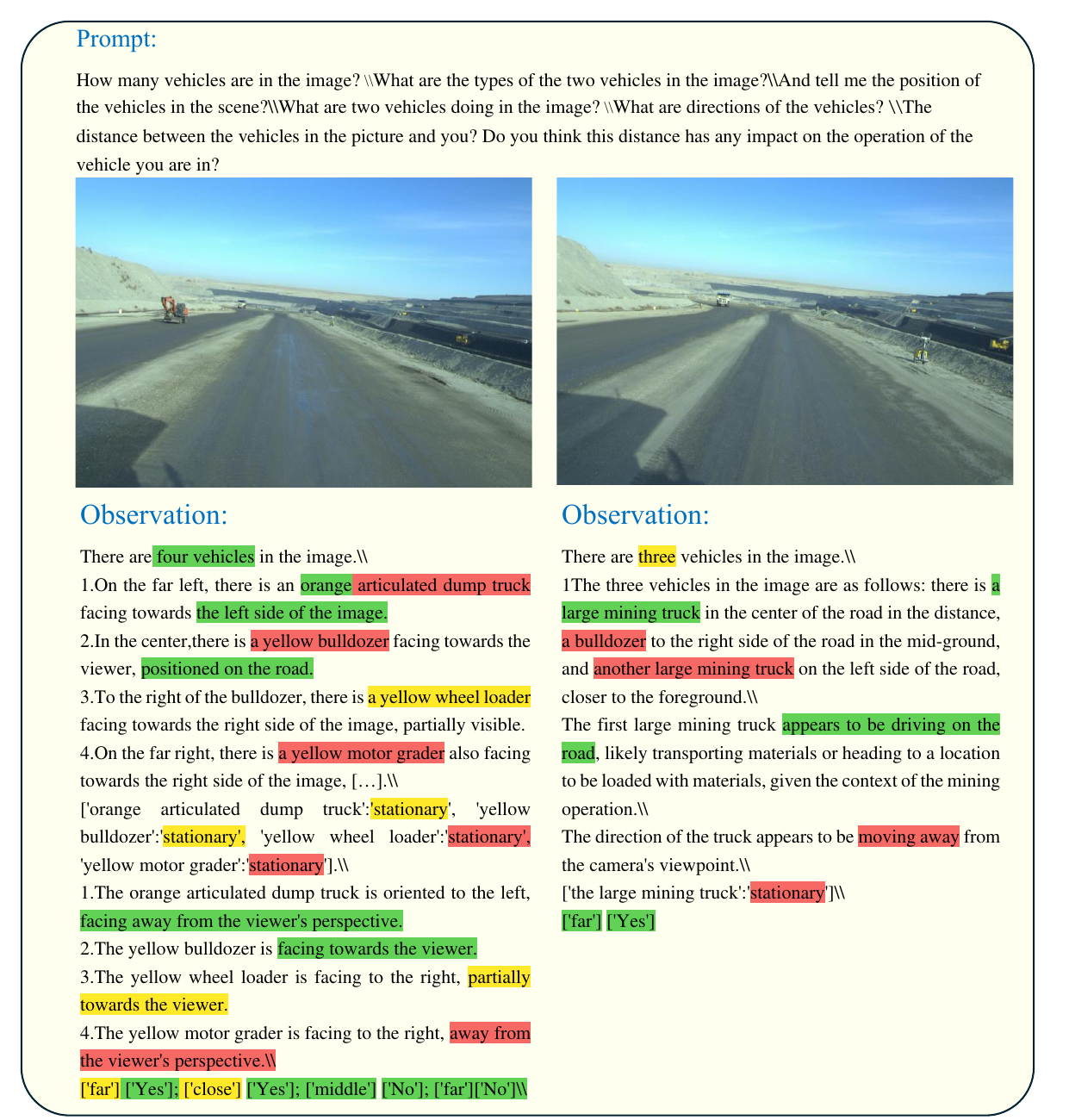}
\caption{\colorbox{green}{Green}highlights the right answer in understanding,\colorbox{red}{Red}highlights the wrong answer in understanding,\colorbox{yellow}{Yellow} highlights the incompetence in performing the task.}
\label{fig_4}
\end{figure*}

\begin{figure*}[h]
\centering
\includegraphics[width=\linewidth]{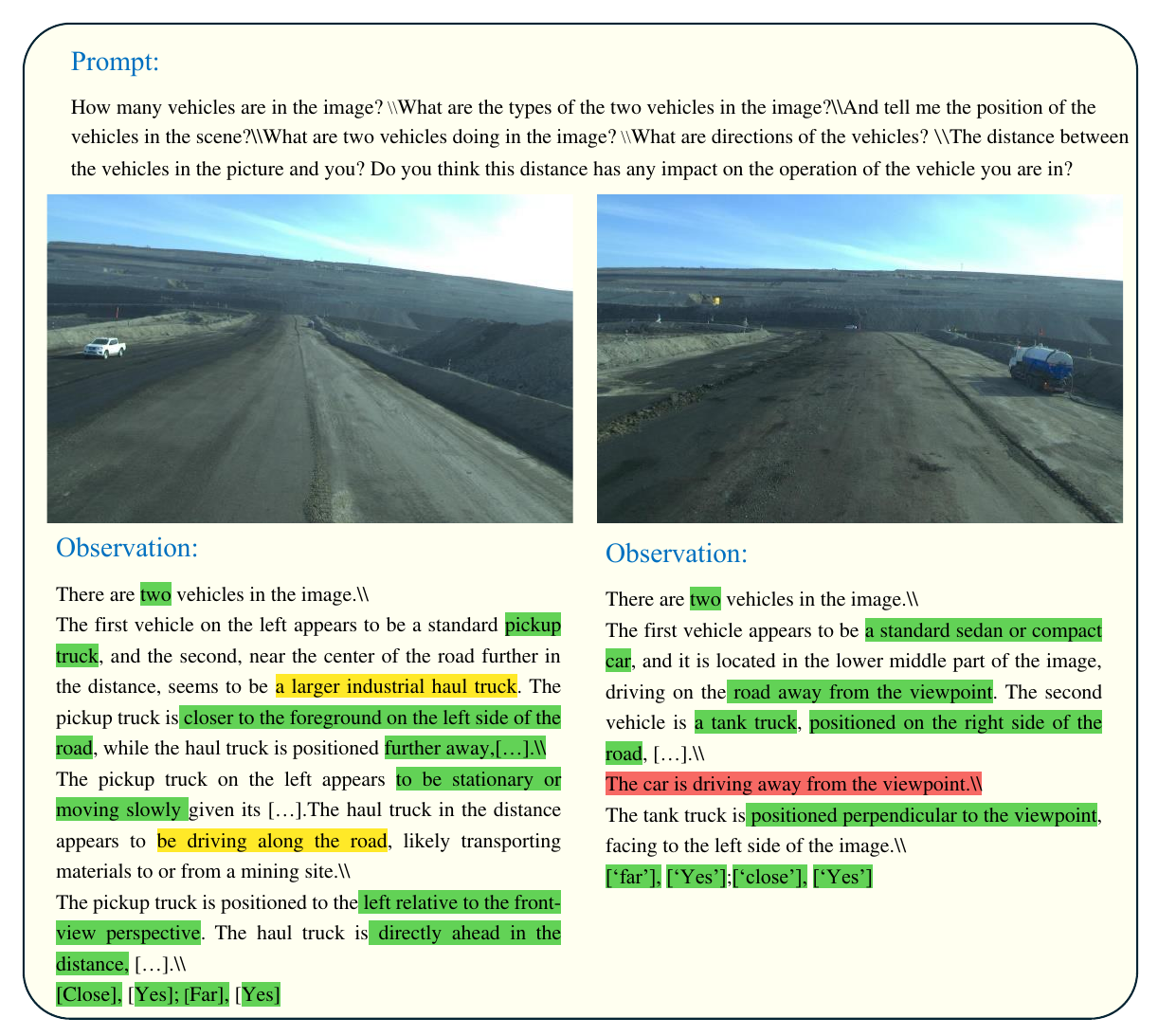}
\caption{\colorbox{green}{Green}highlights the right answer in understanding,\colorbox{red}{Red}highlights the wrong answer in understanding,\colorbox{yellow}{Yellow} highlights the incompetence in performing the task.}
\label{fig_5}
\end{figure*}

\begin{figure*}[!b]
\centering
\includegraphics[width=\linewidth]{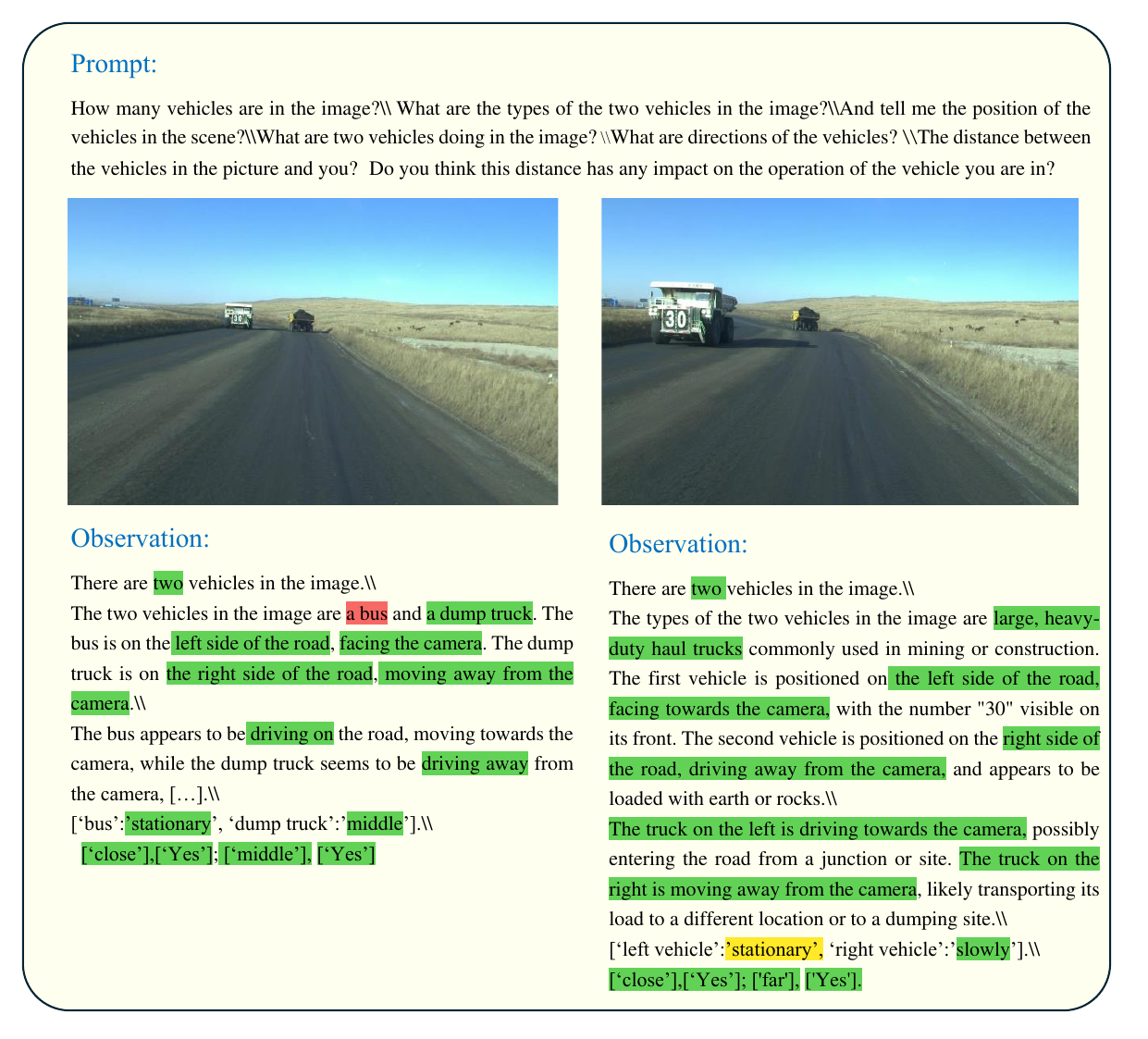}
\caption{\colorbox{green}{Green}highlights the right answer in understanding,\colorbox{red}{Red}highlights the wrong answer in understanding,\colorbox{yellow}{Yellow} highlights the incompetence in performing the task.}
\label{fig_6}
\end{figure*}

\begin{figure*}[h]
\centering
\includegraphics[width=\linewidth]{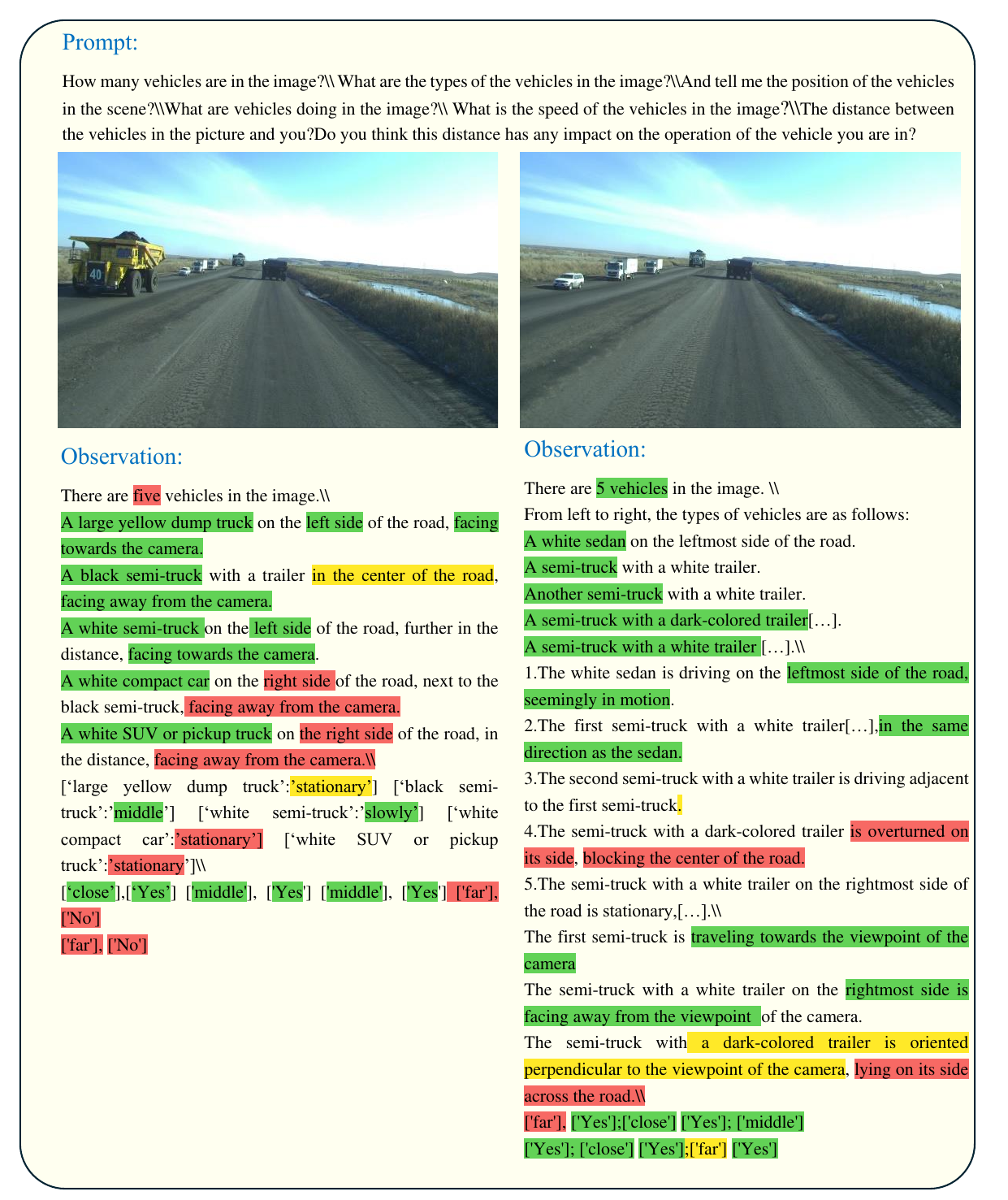}
\caption{\colorbox{green}{Green}highlights the right answer in understanding,\colorbox{red}{Red}highlights the wrong answer in understanding,\colorbox{yellow}{Yellow} highlights the incompetence in performing the task.}
\label{fig_9}
\end{figure*}

Quantity and Type: GPT-4V encountered challenges in accurately identifying vehicle types, influenced by prevalent dust and the considerable distances typical of mining environments. As illustrated in Figures~\ref{fig_2} and \ref{fig_3}, GPT-4V was unable to recognize distant vehicles, such as a black pickup and a yellow truck. Misclassifications were also evident in Figure~\ref{fig_3}, where GPT-4V incorrectly identified a truck. Errors in vehicle counting were apparent, as depicted in Figures~\ref{fig_4} and \ref{fig_5}, where GPT-4V inaccurately counted mining trucks. Additionally, in scenarios like Figure~\ref{fig_9}, GPT-4V either failed to detect cranes or misidentified them, and in instances like Figure~\ref{fig_6}, it detected non-existent vehicles.

\begin{figure*}[h]
\centering
\includegraphics[width=\linewidth]{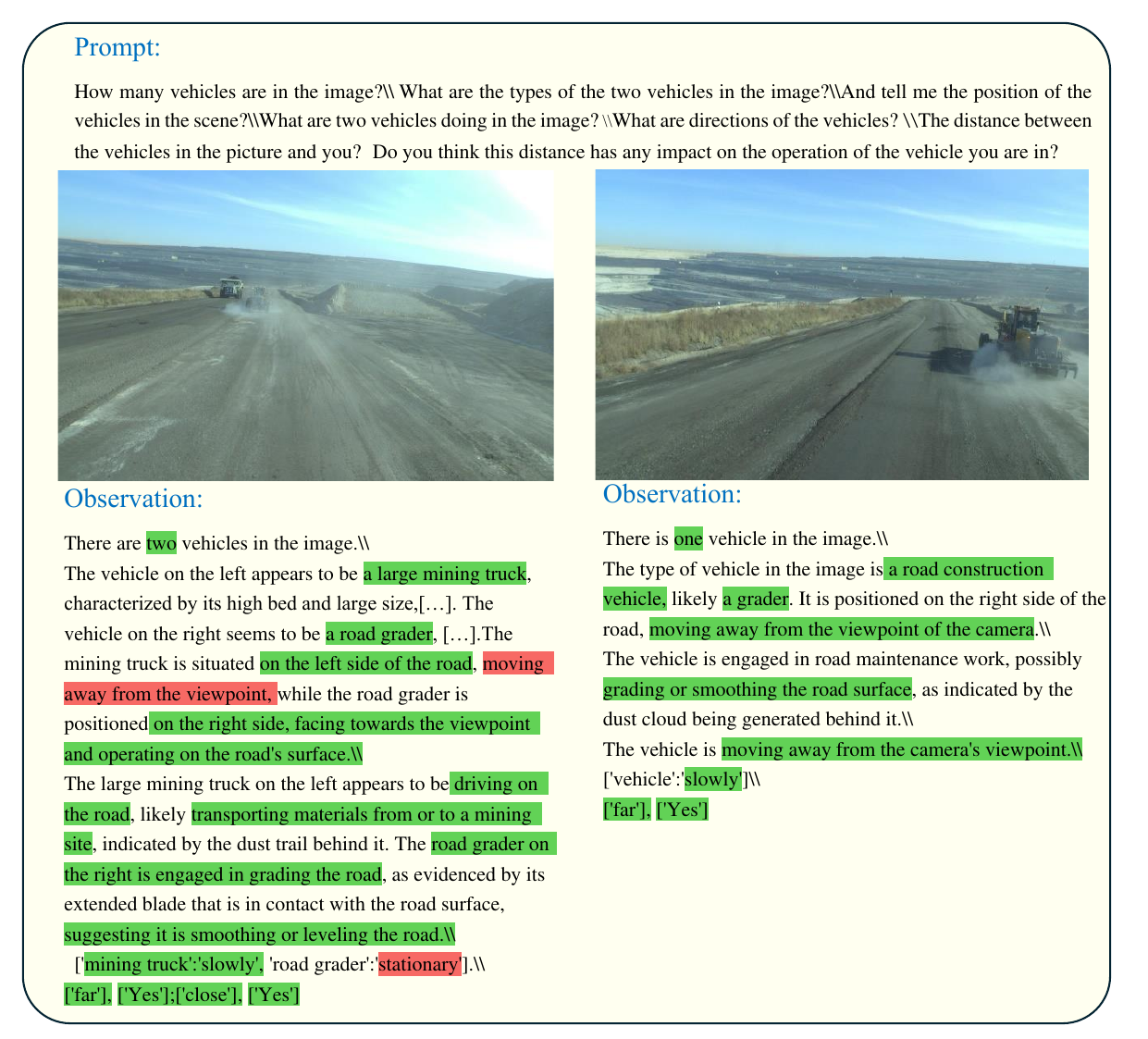}
\caption{\colorbox{green}{Green}highlights the right answer in understanding,\colorbox{red}{Red}highlights the wrong answer in understanding,\colorbox{yellow}{Yellow} highlights the incompetence in performing the task.}
\label{fig_7}
\end{figure*}

\begin{figure*}[h]
\centering
\includegraphics[width=\linewidth]{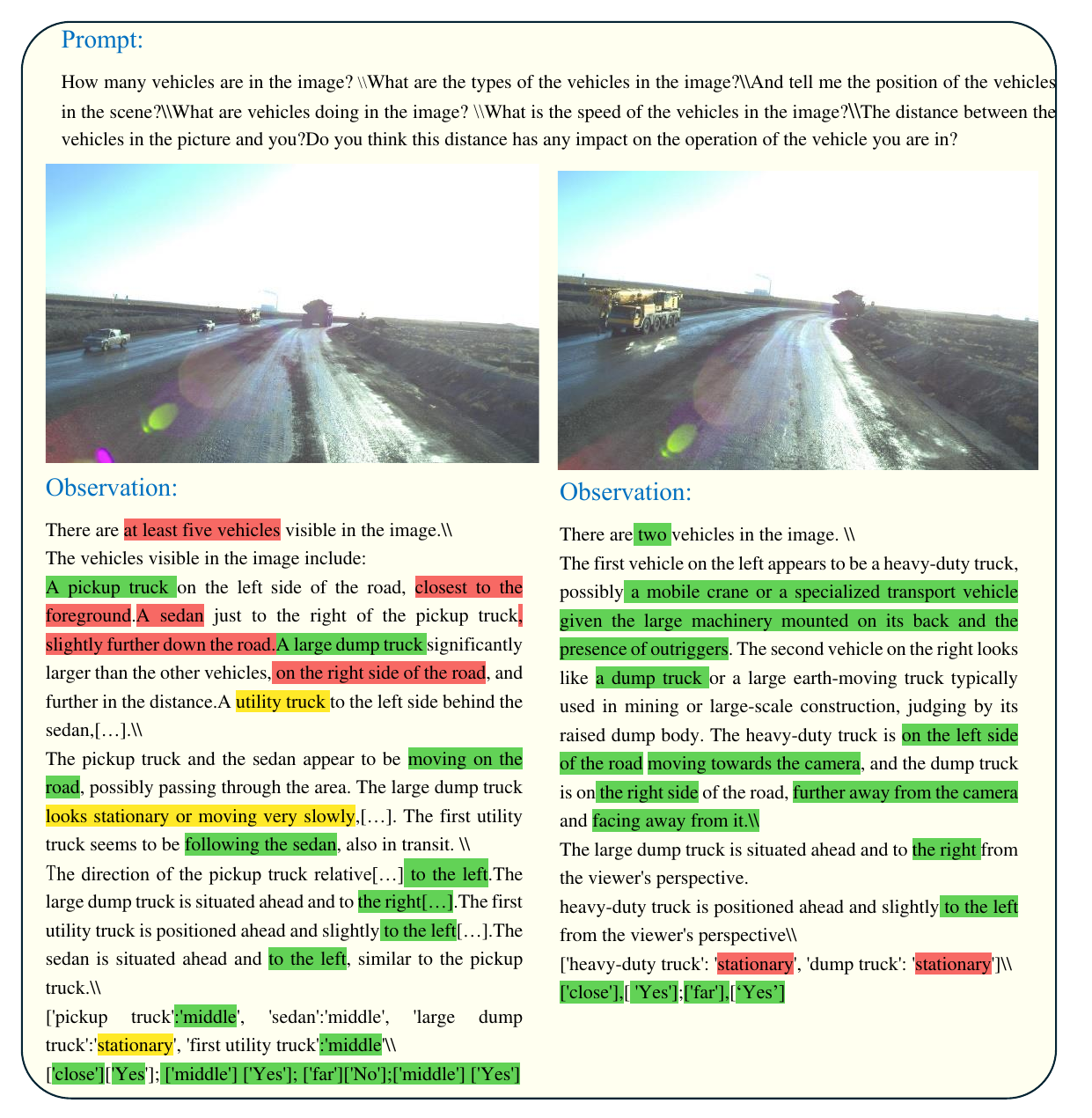}
\caption{\colorbox{green}{Green}highlights the right answer in understanding,\colorbox{red}{Red}highlights the wrong answer in understanding,\colorbox{yellow}{Yellow} highlights the incompetence in performing the task.}
\label{fig_10}
\end{figure*}

Speed and Direction: GPT-4V was unable to provide speed and direction data for some vehicles because GPT-4V couldn't detect the vehicles that are too far, as observed in Figure~\ref{fig_2} and Figure~\ref{fig_3}. As the assessments involved static images, GPT-4V inaccurately perceived all vehicles as stationary, including mining trucks in motion in Figure~\ref{fig_4} and Figure~\ref{fig_7}, and a moving crane in Figure~\ref{fig_10}. In fact,  based on the surrounding environment, it is possible to determine that the vehicle is in motion. Generally, GPT-4V used visible dust trails to gauge vehicle speed, and while most of its speed and direction judgments were accurate, discrepancies remained.

\begin{figure*}[h]
\centering
\includegraphics[width=\linewidth]{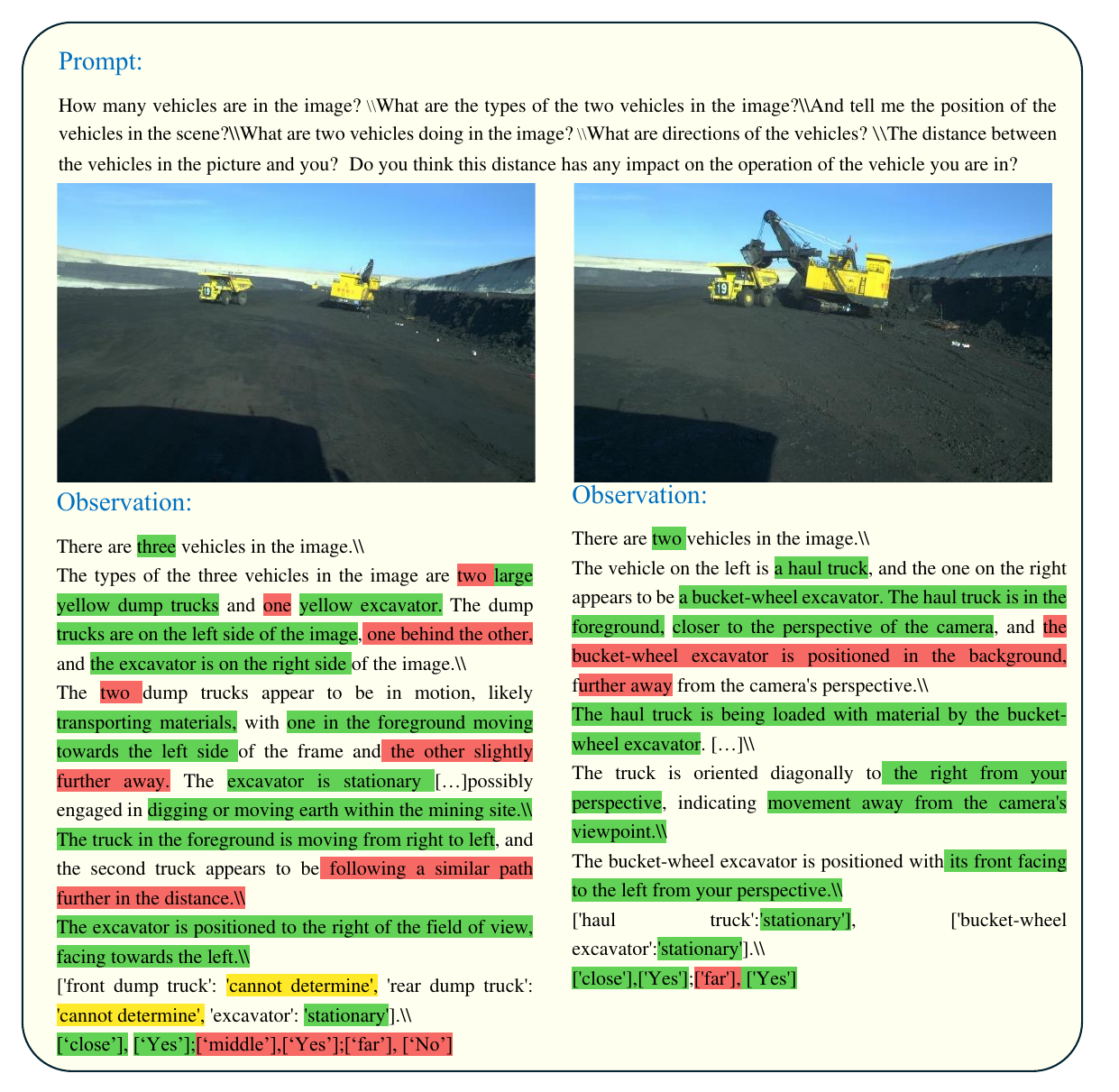}
\caption{\colorbox{green}{Green}highlights the right answer in understanding,\colorbox{red}{Red}highlights the wrong answer in understanding,\colorbox{yellow}{Yellow} highlights the incompetence in performing the task.}
\label{fig_8}
\end{figure*}

Status: GPT-4V frequently mispredicted the operational status of specialized vehicles such as mining trucks and excavators. For instance, it inaccurately concluded that excavators were engaged in digging activities when they were merely relocating, as depicted in Figures~\ref{fig_2} and \ref{fig_3}. Similar errors were observed in Figure~\ref{fig_10}, where mining trucks were mistakenly presumed to be involved in construction tasks. In contrast, GPT-4V correctly identified the activities of a water truck in Figure~\ref{fig_5} and excavators in Figure~\ref{fig_8}. Overall, GPT-4V demonstrates a lack of familiarity with the behaviors and operational dynamics of mining trucks and machinery.

Relative Distance: GPT-4V accurately assessed the positions of identified vehicles relative to the observer's vehicle. For instance, it maintained a safe distance from a sedan in Figure~\ref{fig_9} and precisely determined the orientation of an excavator and a mining truck in Figure~\ref{fig_4} In Figure~\ref{fig_5}, it accurately judged the direction of a pickup and the distance to a distant truck, demonstrating precision in controlling the direction and distance of vehicles relative to itself.

This study highlights GPT-4V’s strengths and weaknesses in vehicle scene understanding within mining environments, underscoring the need for further refinement to enhance its performance in these complex settings.

\subsection{Understanding of Rail Pile in Scenes}
\begin{figure*}[h]
\centering
\includegraphics[width=\linewidth]{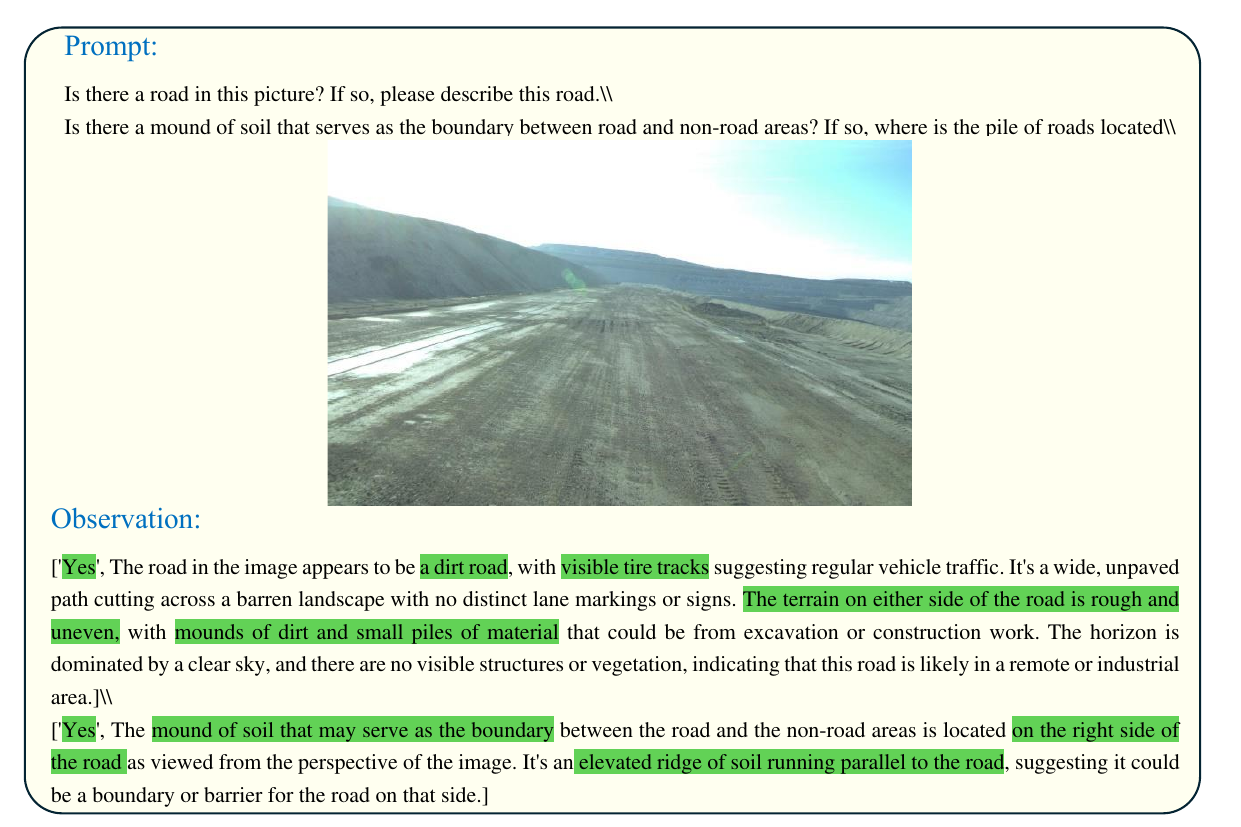}
\caption{\colorbox{green}{Green}highlights the right answer in understanding,\colorbox{red}{Red}highlights the wrong answer in understanding,\colorbox{yellow}{Yellow} highlights the incompetence in performing the task.}
\label{fig_11}
\end{figure*}

\begin{figure*}[h]
\centering
\includegraphics[width=\linewidth]{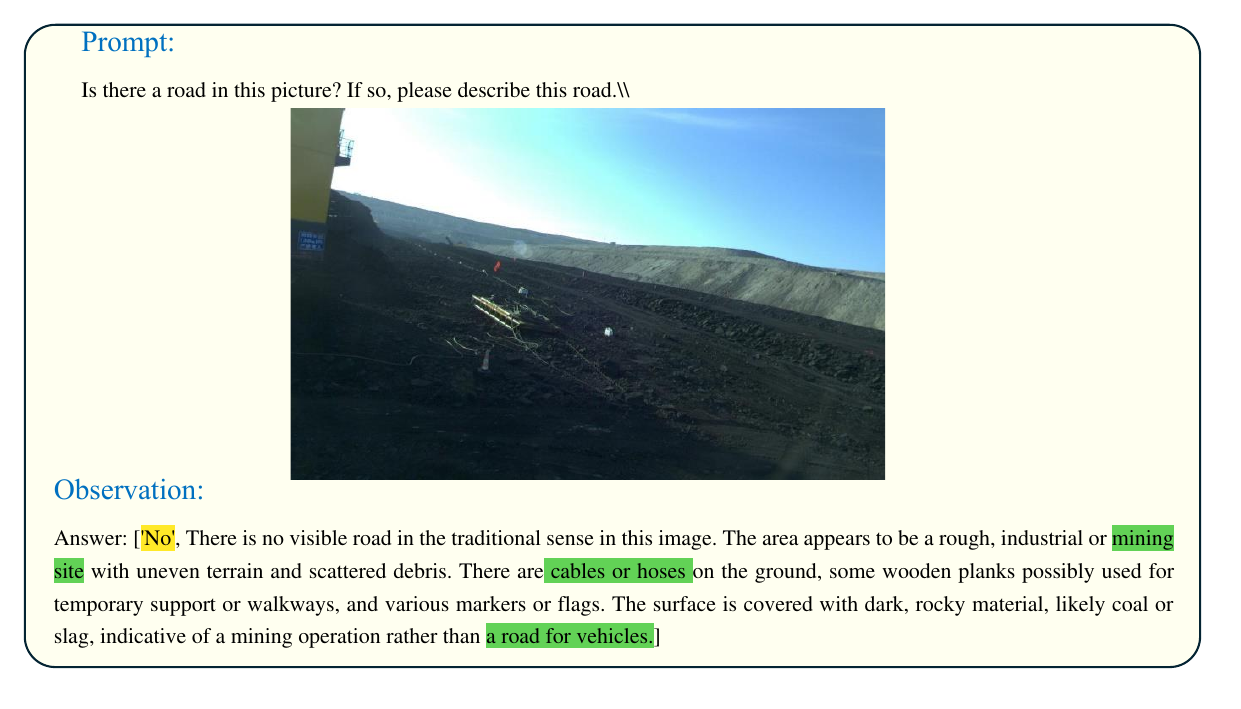}
\caption{\colorbox{green}{Green}highlights the right answer in understanding,\colorbox{red}{Red}highlights the wrong answer in understanding,\colorbox{yellow}{Yellow} highlights the incompetence in performing the task.}
\label{fig_12}
\end{figure*}

\begin{figure*}[h]
\centering
\includegraphics[width=\linewidth]{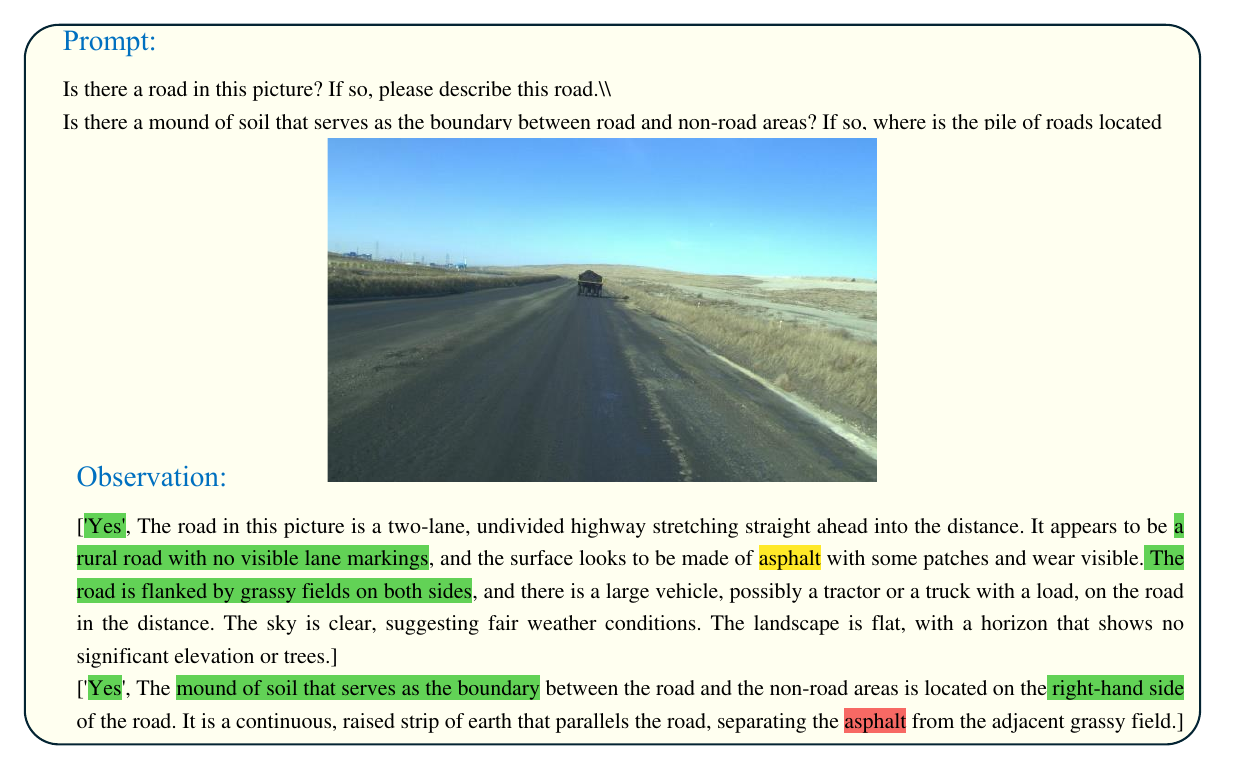}
\caption{\colorbox{green}{Green}highlights the right answer in understanding,\colorbox{red}{Red}highlights the wrong answer in understanding,\colorbox{yellow}{Yellow} highlights the incompetence in performing the task.}
\label{fig_13}
\end{figure*}

\begin{figure*}[h]
\centering
\includegraphics[width=\linewidth]{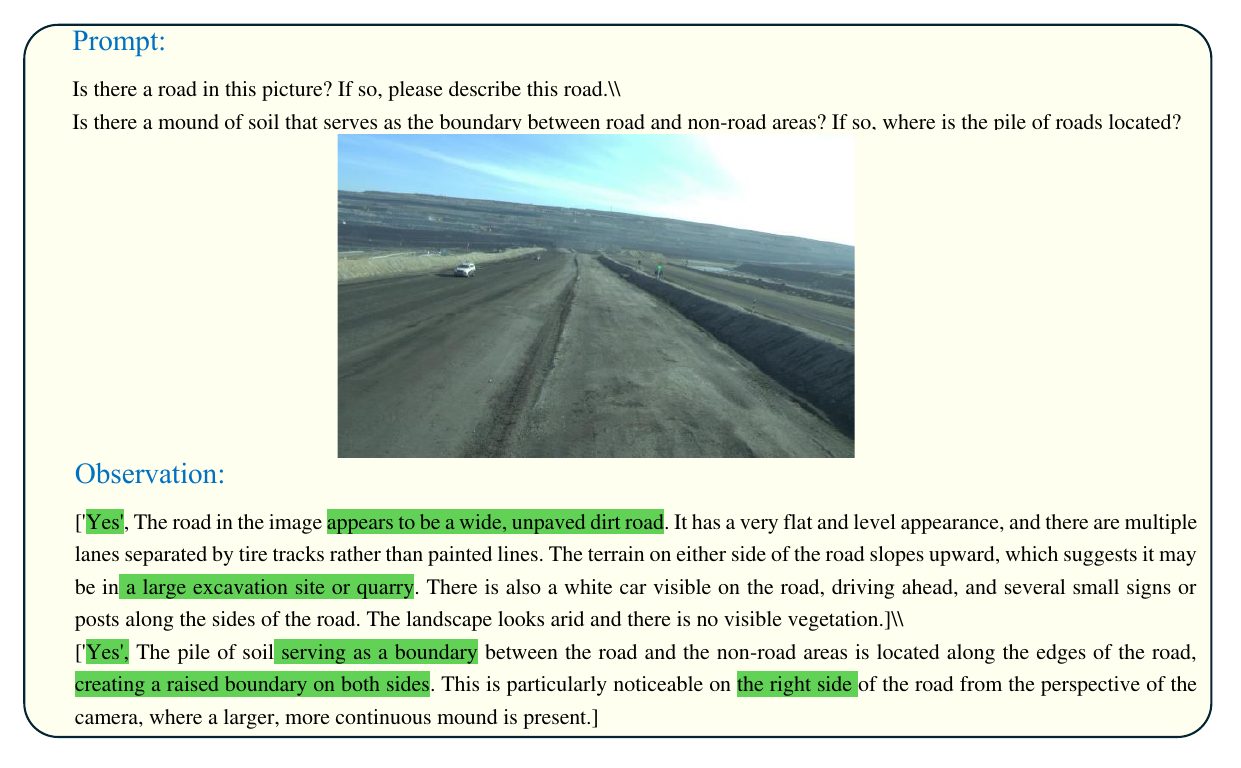}
\caption{\colorbox{green}{Green}highlights the right answer in understanding,\colorbox{red}{Red}highlights the wrong answer in understanding,\colorbox{yellow}{Yellow} highlights the incompetence in performing the task.}
\label{fig_14}
\end{figure*}

In our evaluation of GPT-4V, we investigated its capability to comprehend unstructured roads and ore piles within mining environments, characterized by dynamically changing landscapes and obscure boundaries. GPT-4V demonstrated proficiency in scenarios with clearer road features, accurately identifying ore piles along roads as boundaries, exemplified by images Figure~\ref{fig_11} and Figure~\ref{fig_14}.Conversely, in more complex segments of mining areas, such as depicted in Figure~\ref{fig_12}. GPT-4V recognized mining slag but failed to identify the ore piles that accumulated from it.
Although GPT-4V effectively discerned ore piles adjacent to roads and accurately identified road boundaries, it encountered difficulties in consistently recognizing the material composition of road surfaces. For instance, in Figure~\ref{fig_13}, it erroneously classified a dirt road as asphalt. Additionally, in mining areas where road boundaries are defined by vegetation rather than ore piles, GPT-4V successfully adapted and recognized these natural boundaries. This adaptability highlights GPT-4V's capacity to interpret various environmental cues within the challenging contexts of mining areas. The results demonstrate that GPT-4V has a robust understanding of mine roads, enabling it to accurately identify roads in these areas.

\subsection{Understanding of mechanical infrastructure in Scenes}
\begin{figure*}[h]
\centering
\includegraphics[width=\linewidth]{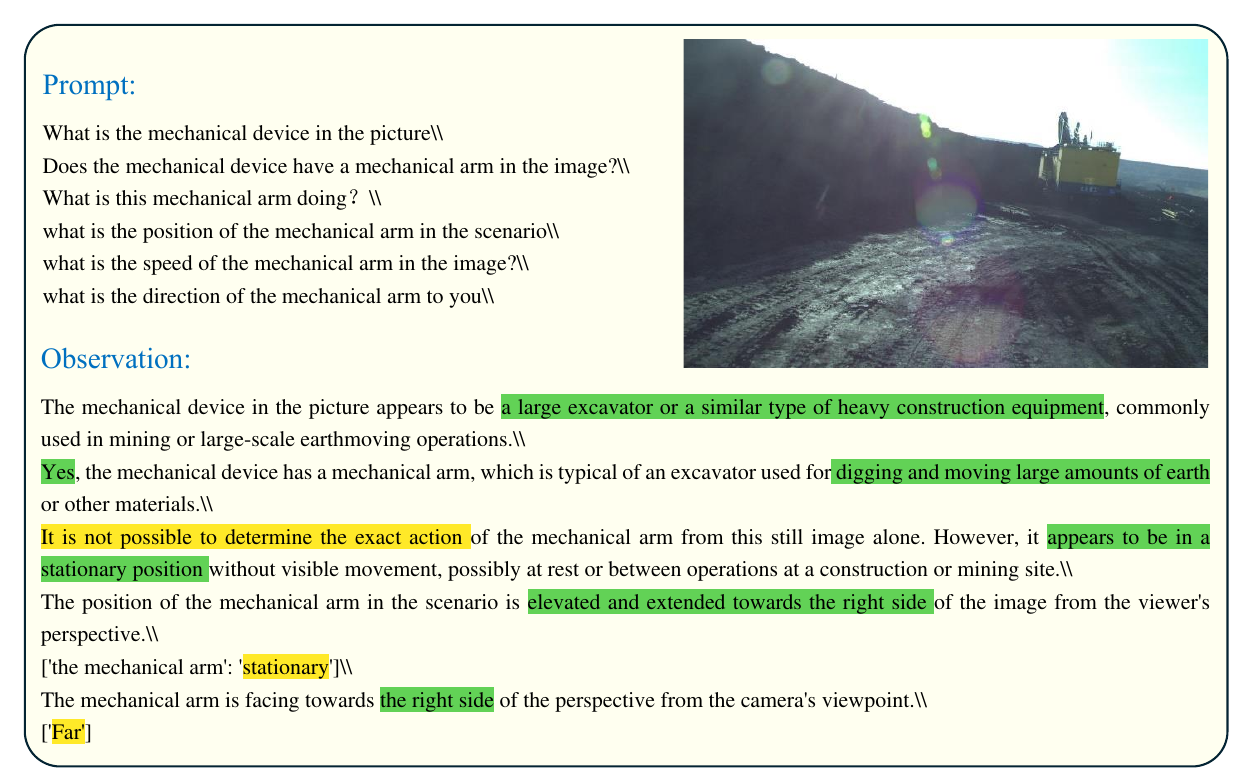}
\caption{\colorbox{green}{Green}highlights the right answer in understanding,\colorbox{red}{Red}highlights the wrong answer in understanding,\colorbox{yellow}{Yellow} highlights the incompetence in performing the task.}
\label{fig_16}
\end{figure*}

\begin{figure*}[h]
\centering
\includegraphics[width=\linewidth]{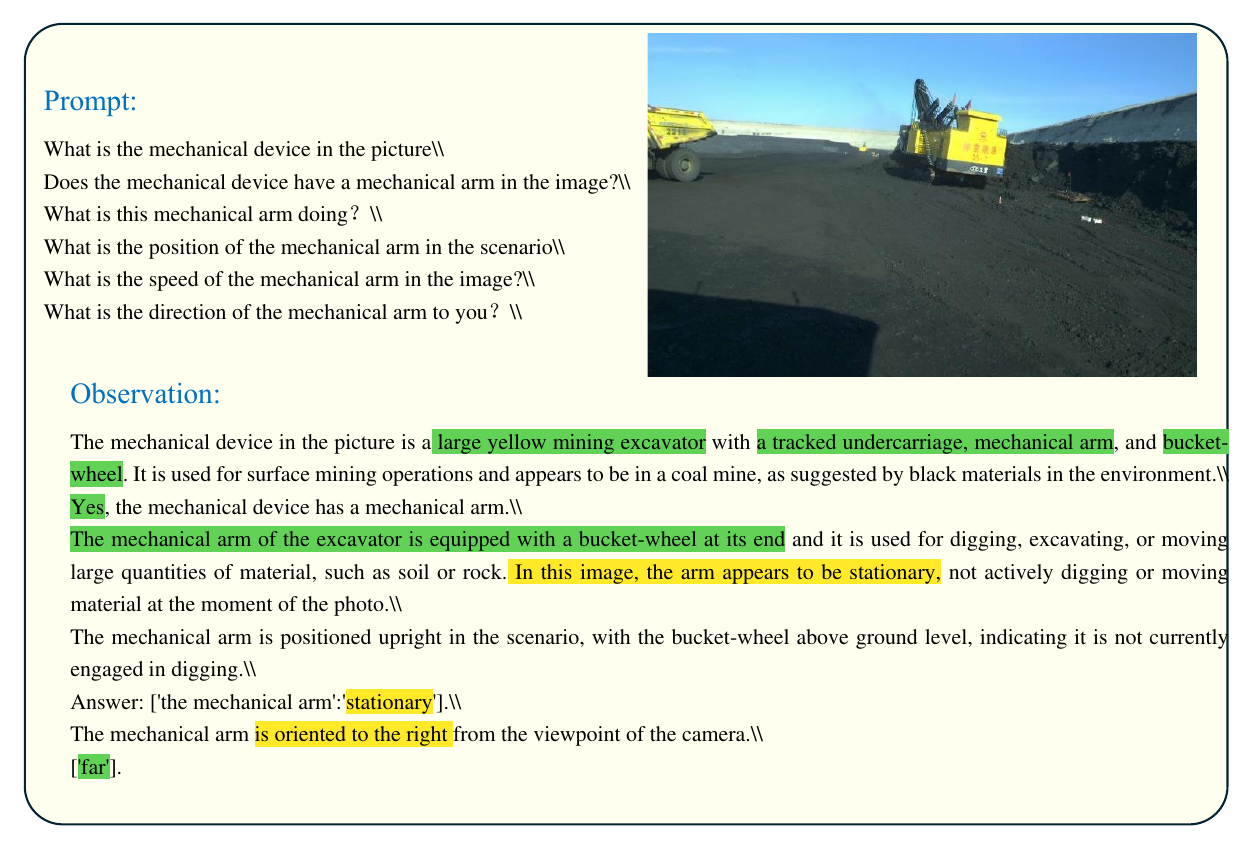}
\caption{\colorbox{green}{Green}highlights the right answer in understanding,\colorbox{red}{Red}highlights the wrong answer in understanding,\colorbox{yellow}{Yellow} highlights the incompetence in performing the task.}
\label{fig_17}
\end{figure*}

\begin{figure*}[h]
\centering
\includegraphics[width=\linewidth]{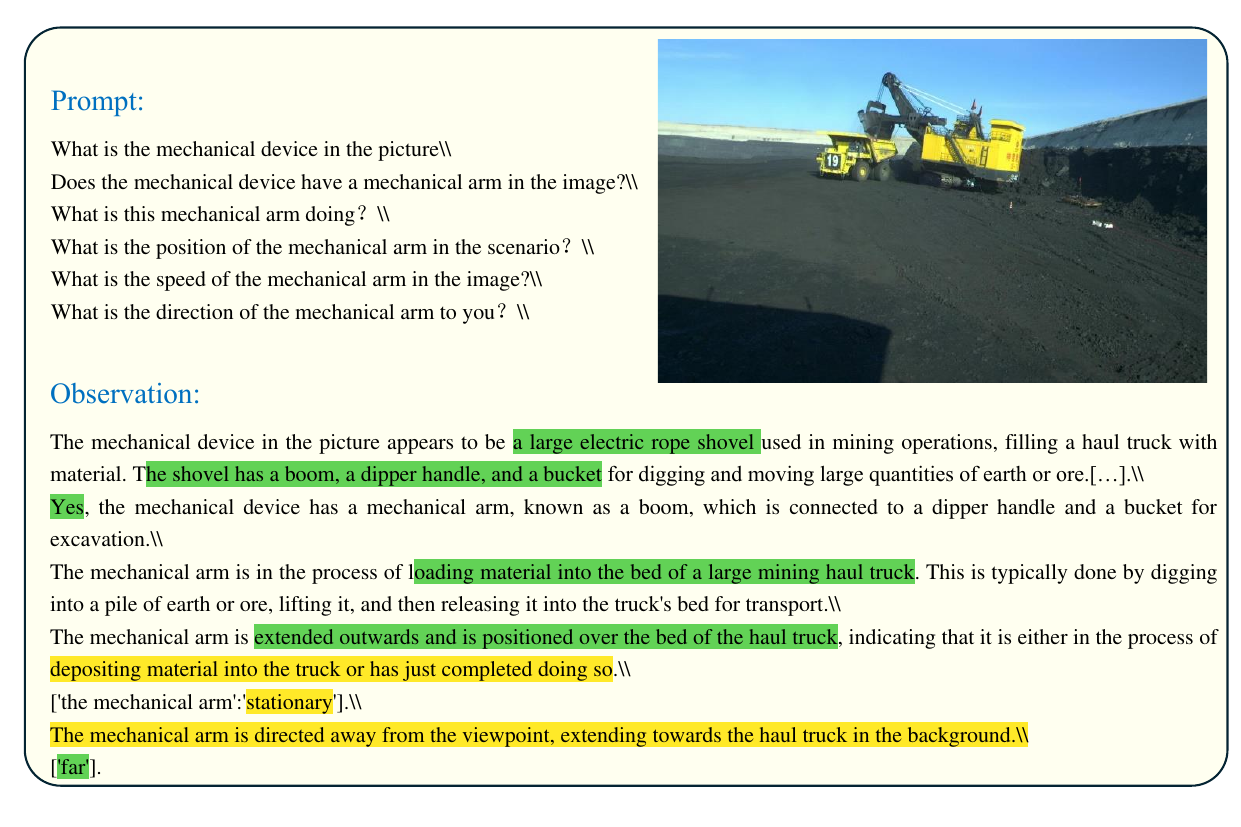}
\caption{\colorbox{green}{Green}highlights the right answer in understanding,\colorbox{red}{Red}highlights the wrong answer in understanding,\colorbox{yellow}{Yellow} highlights the incompetence in performing the task.}
\label{fig_18}
\end{figure*}

\begin{figure*}[h]
\centering
\includegraphics[width=\linewidth]{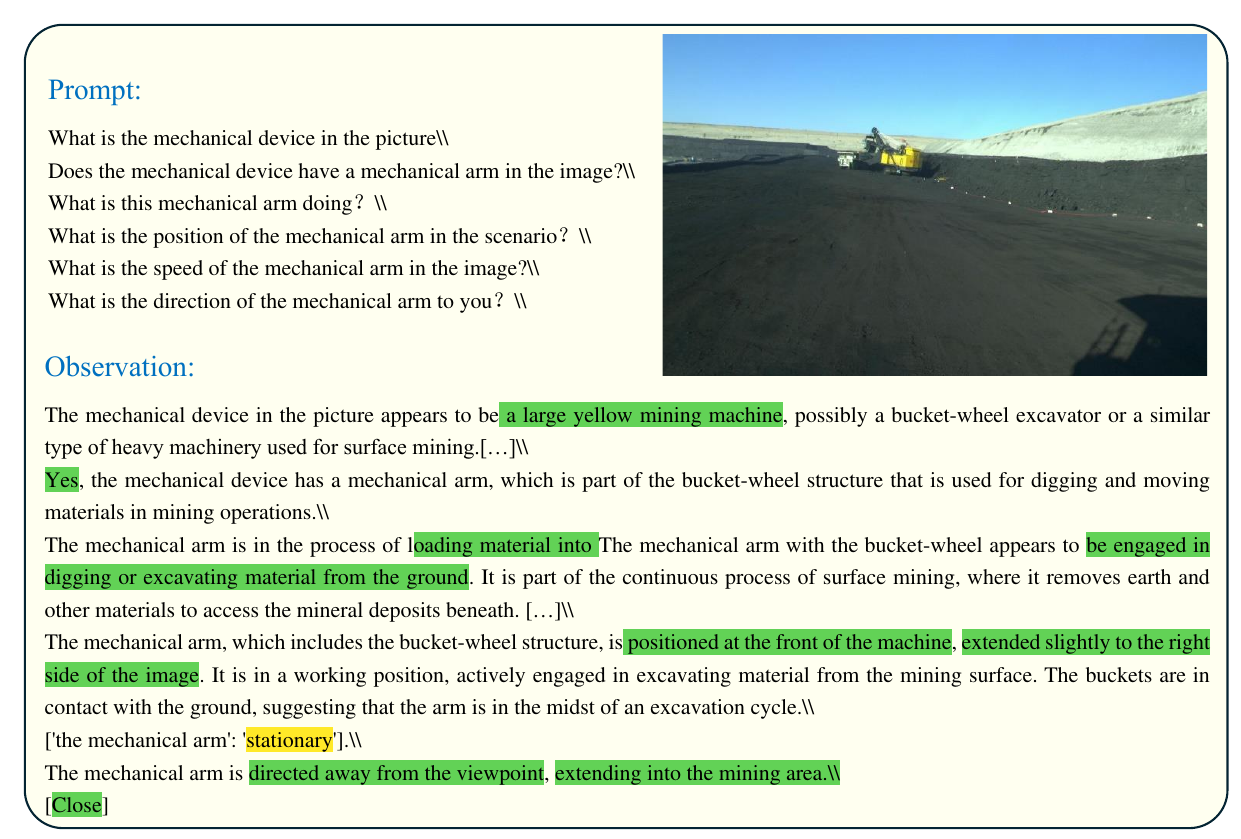}
\caption{\colorbox{green}{Green}highlights the right answer in understanding,\colorbox{red}{Red}highlights the wrong answer in understanding,\colorbox{yellow}{Yellow} highlights the incompetence in performing the task.}
\label{fig_19}
\end{figure*}

\begin{figure*}[h]
\centering
\includegraphics[width=\linewidth]{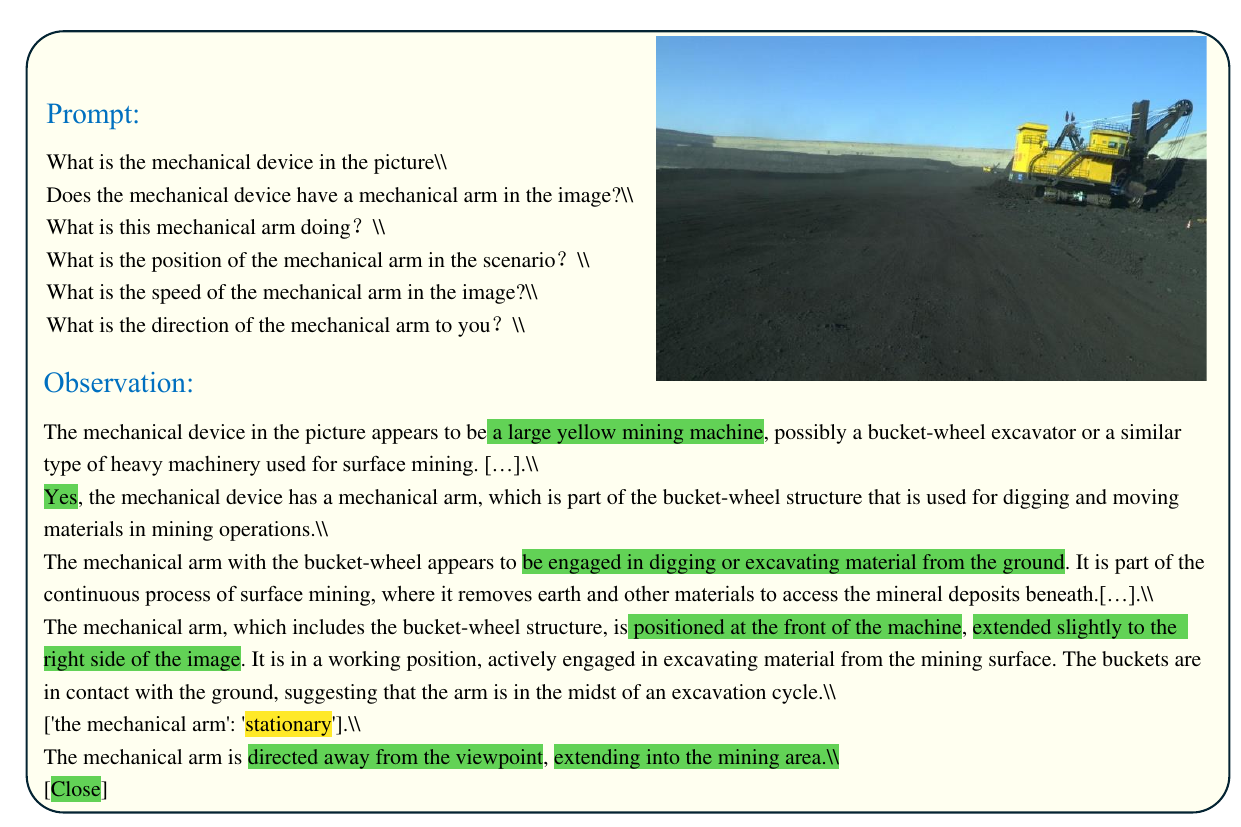}
\caption{\colorbox{green}{Green}highlights the right answer in understanding,\colorbox{red}{Red}highlights the wrong answer in understanding,\colorbox{yellow}{Yellow} highlights the incompetence in performing the task.}
\label{fig_20}
\end{figure*}

In mining environments, recognizing extensive mechanical infrastructure is essential, particularly excavators which include mechanical arms. The operational status of these excavators and the positioning of their arms significantly impact vehicle maneuverability. When these arms are active and rotating, vehicles must maneuver carefully to avoid them. While GPT-4V effectively identifies excavators and their mechanical arms, its ability to accurately assess the movement of these arms requires enhancement. For example, GPT-4V's evaluation of the arm's motion was found lacking in Figure~\ref{fig_16}. Additionally, GPT-4V struggled with accurately predicting the position and directional movement of the arms, as evidenced in Figure~\ref{fig_17}, Figure~\ref{fig_18} and Figure~\ref{fig_19}. GPT-4V demonstrated excellent performance specifically about mechanical arms in Figure~\ref{fig_20}. Since these are single images, GPT-4V cannot accurately assess the motion state of the mechanical arm. In our test, GPT-4V has excellent capabilities in understanding mechanical arms.

\subsection{Understanding of traffic traffic signs in Scenes}
\begin{figure*}[h]
\centering
\includegraphics[width=\linewidth]{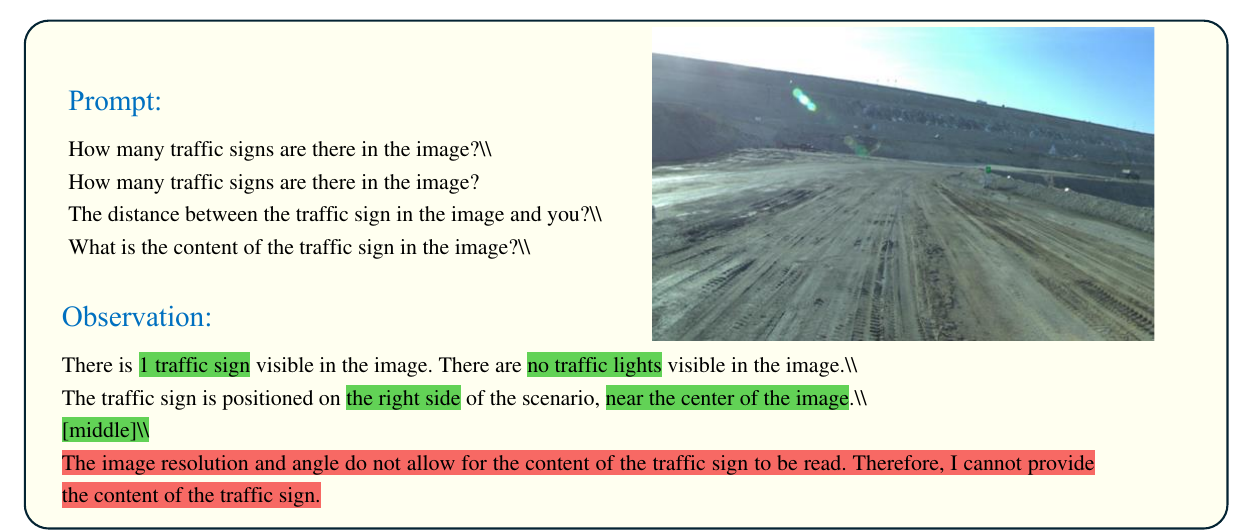}
\caption{\colorbox{green}{Green}highlights the right answer in understanding,\colorbox{red}{Red}highlights the wrong answer in understanding,\colorbox{yellow}{Yellow} highlights the incompetence in performing the task.}
\label{fig_21}
\end{figure*}

\begin{figure*}[h]
\centering
\includegraphics[width=\linewidth]{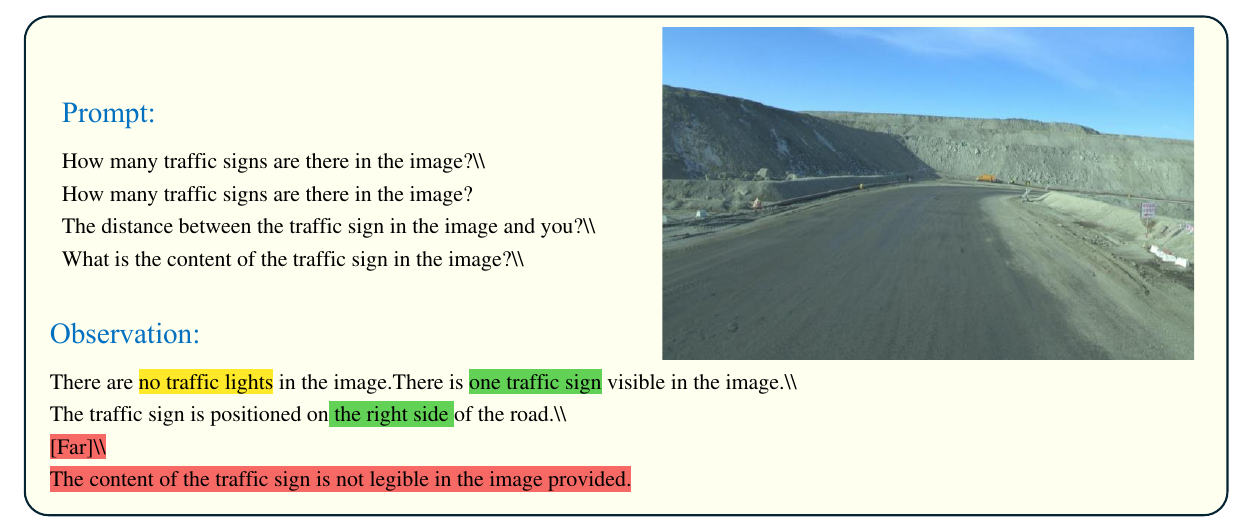}
\caption{\colorbox{green}{Green}highlights the right answer in understanding,\colorbox{red}{Red}highlights the wrong answer in understanding,\colorbox{yellow}{Yellow} highlights the incompetence in performing the task.}
\label{fig_22}
\end{figure*}

\begin{figure*}[h]
\centering
\includegraphics[width=\linewidth]{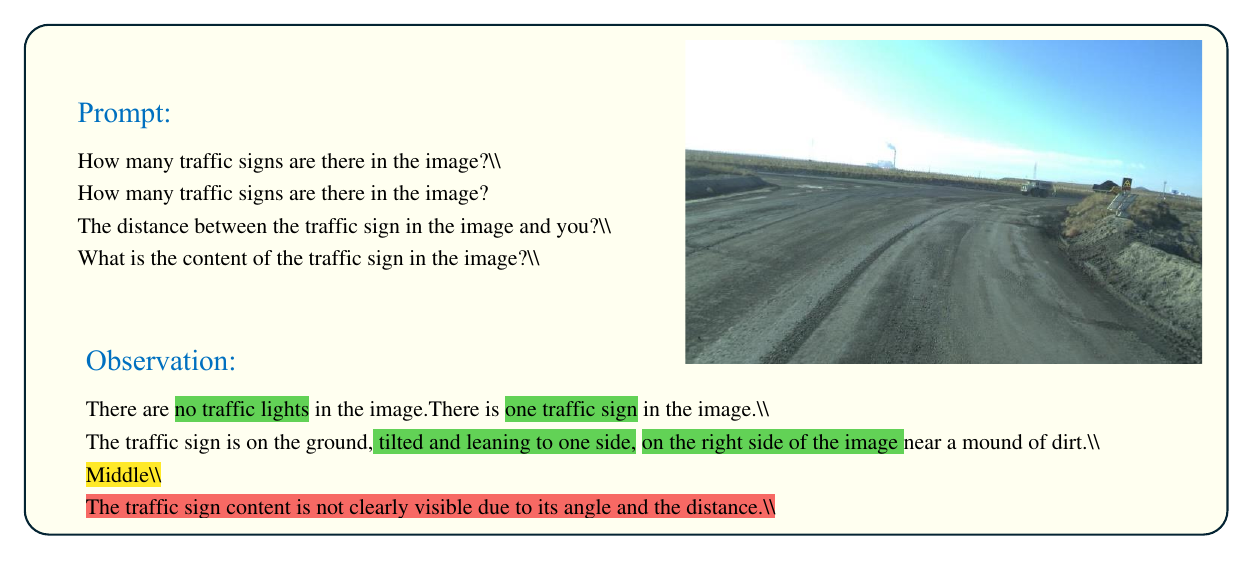}
\caption{\colorbox{green}{Green}highlights the right answer in understanding,\colorbox{red}{Red}highlights the wrong answer in understanding,\colorbox{yellow}{Yellow} highlights the incompetence in performing the task.}
\label{fig_27}
\end{figure*}

\begin{figure*}[h]
\centering
\includegraphics[width=\linewidth]{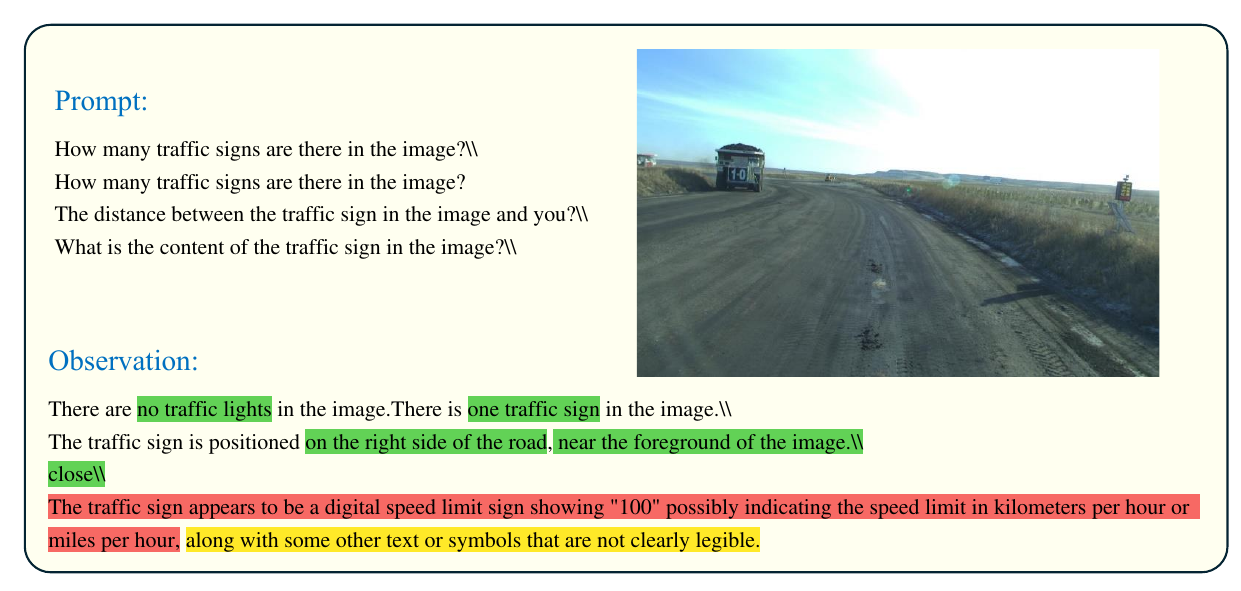}
\caption{\colorbox{green}{Green}highlights the right answer in understanding,\colorbox{red}{Red}highlights the wrong answer in understanding,\colorbox{yellow}{Yellow} highlights the incompetence in performing the task.}
\label{fig_29}
\end{figure*}

\begin{figure*}[h]
\centering
\includegraphics[width=\linewidth]{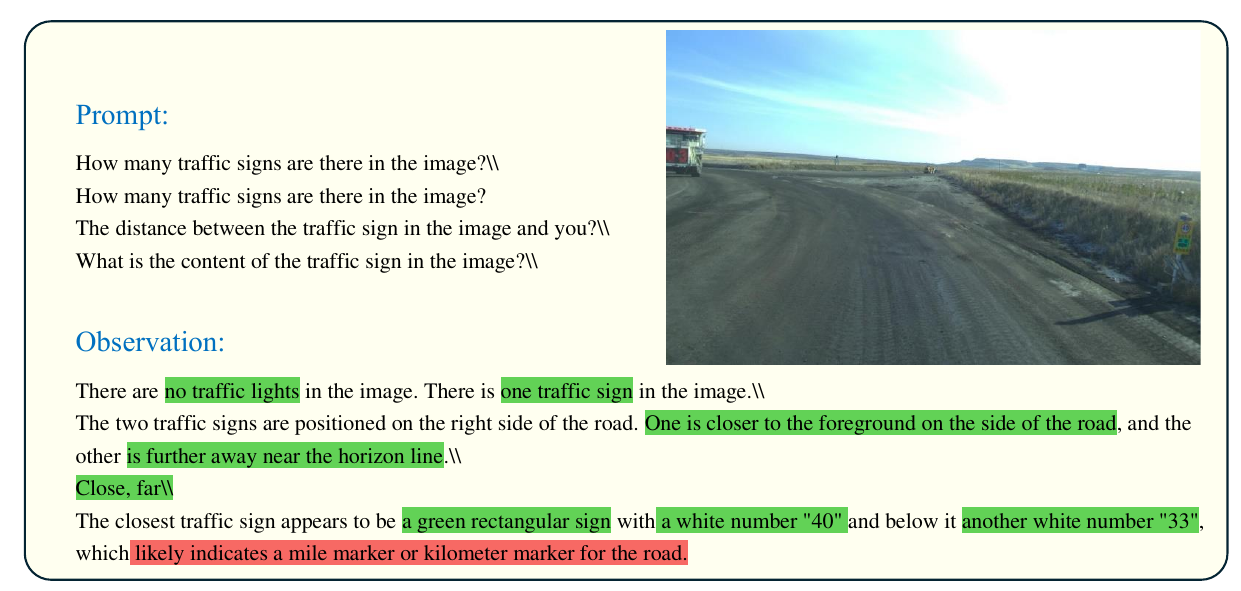}
\caption{\colorbox{green}{Green}highlights the right answer in understanding,\colorbox{red}{Red}highlights the wrong answer in understanding,\colorbox{yellow}{Yellow} highlights the incompetence in performing the task.}
\label{fig_30}
\end{figure*}

\begin{figure*}[h]
\centering
\includegraphics[width=\linewidth]{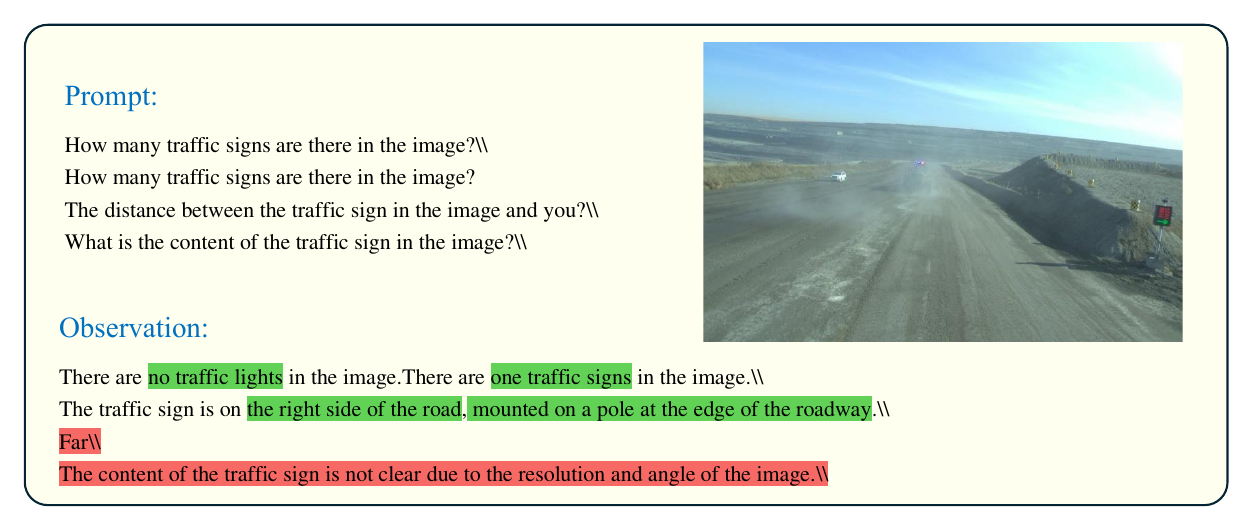}
\caption{\colorbox{green}{Green}highlights the right answer in understanding,\colorbox{red}{Red}highlights the wrong answer in understanding,\colorbox{yellow}{Yellow} highlights the incompetence in performing the task.}
\label{fig_32}
\end{figure*}

\begin{figure*}[h]
\centering
\includegraphics[width=\linewidth]{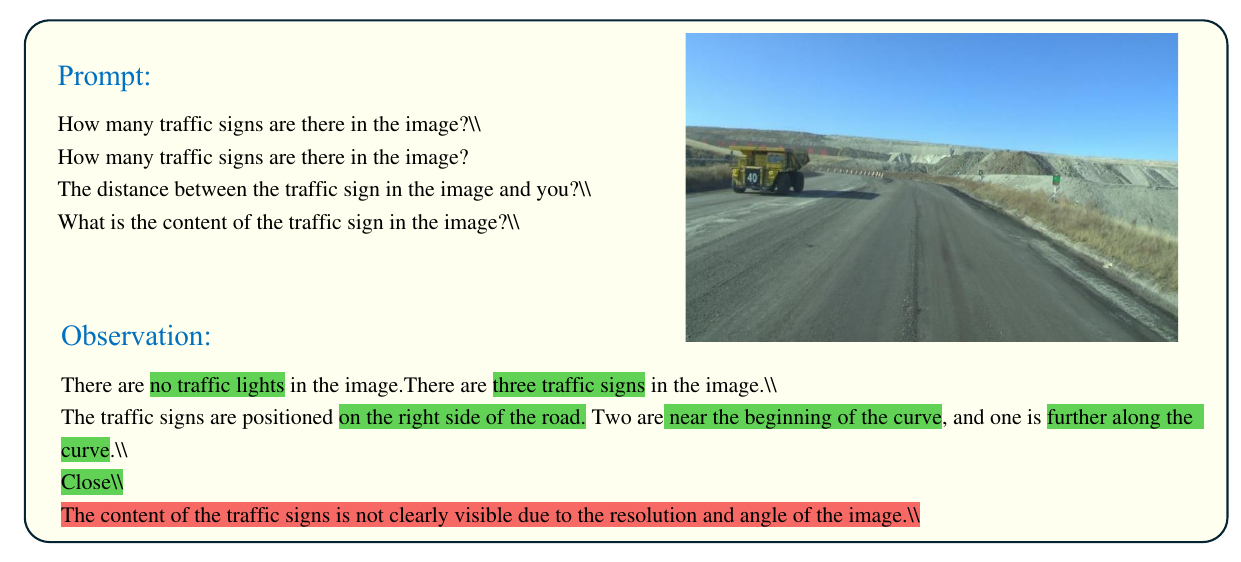}
\caption{\colorbox{green}{Green}highlights the right answer in understanding,\colorbox{red}{Red}highlights the wrong answer in understanding,\colorbox{yellow}{Yellow} highlights the incompetence in performing the task.}
\label{fig_33}
\end{figure*}

\begin{figure*}[h]
\centering
\includegraphics[width=\linewidth]{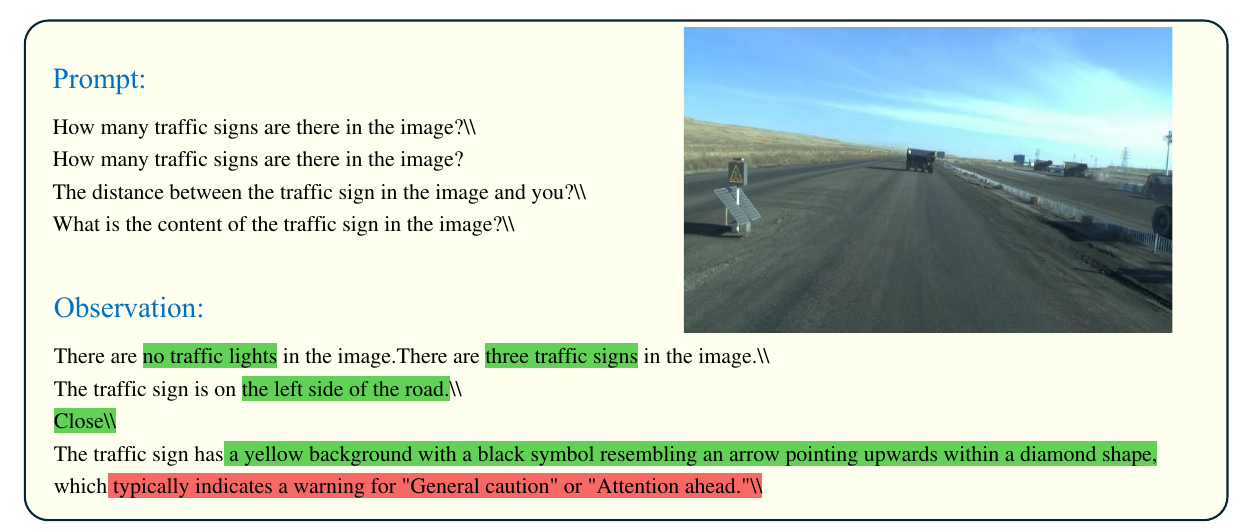}
\caption{\colorbox{green}{Green}highlights the right answer in understanding,\colorbox{red}{Red}highlights the wrong answer in understanding,\colorbox{yellow}{Yellow} highlights the incompetence in performing the task.}
\label{fig_34}
\end{figure*}

\begin{figure*}[h]
\centering
\includegraphics[width=\linewidth]{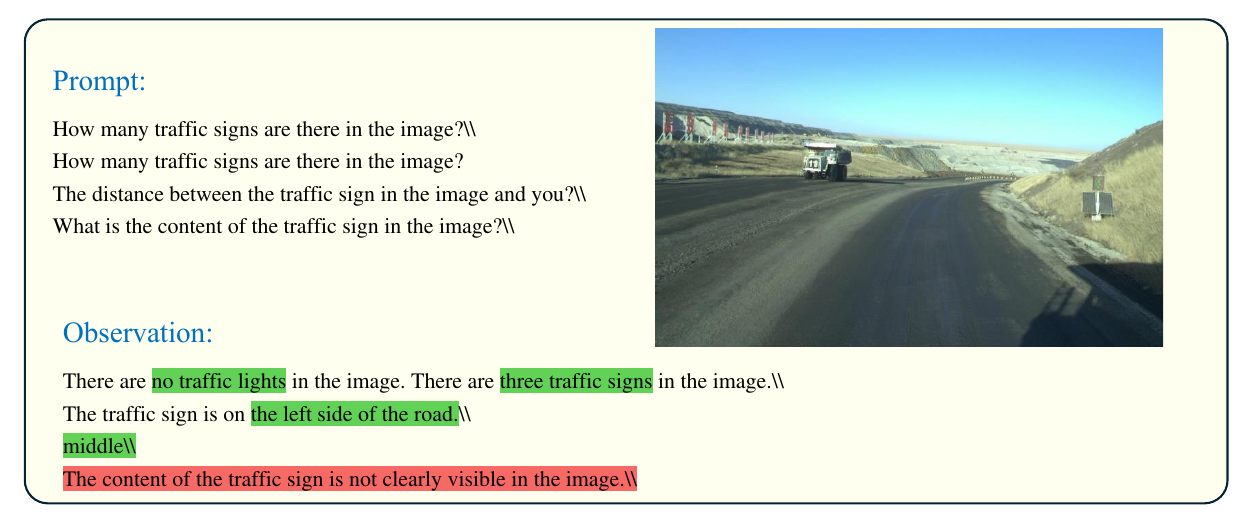}
\caption{\colorbox{green}{Green}highlights the right answer in understanding,\colorbox{red}{Red}highlights the wrong answer in understanding,\colorbox{yellow}{Yellow} highlights the incompetence in performing the task.}
\label{fig_35}
\end{figure*}

Traffic signals and signage are crucial for vehicle navigation in mining areas, as they guide driving behaviors such as speed and turning based on their recognized contents. However, the ability of GPT-4V to accurately and swiftly recognize the contents of these traffic signals and signs remains a concern. We conducted tests to assess GPT-4V's accuracy in identifying the number, position, distance, and content of traffic signals and signs adjacent to roadways. GPT-4V demonstrated reasonably accurate estimations of the number, position, and distance of traffic signals and signs. 
However, frequent errors were noted in content recognition. Notably, GPT-4V struggled with deciphering the content of signs. In Figure~\ref{fig_21}, the traffic sign signaling a left turn was not recognized by GPT-4V, possibly due to the excessive distance making the content unclear. A similar issue is depicted in Figure~\ref{fig_22}. GPT-4V also had difficulty recognizing less common traffic signs, such as the T-junction traffic sign in Figure~\ref{fig_27}, the right-turning traffic sign in Figure~\ref{fig_33}, the keep distance traffic sign in Figure~\ref{fig_34}, and the no overtaking traffic sign in Figure~\ref{fig_35}. Although GPT-4V could recognize numbers on traffic signs as shown in Figure~\ref{fig_30}, it did not understand their significance; "40" denotes the maximum speed limit, and "33" indicates the current speed of the vehicle. Additionally, GPT-4V failed to identify Chinese traffic signs, as seen in Figure~\ref{fig_32}. In Figure~\ref{fig_29}, GPT-4V not only failed to recognize Chinese characters but also erroneously identified them as numbers.
These errors can be fatal for vehicles in motion. Traffic signs play a critical role by providing essential alerts and directing vehicular navigation. However, GPT-4V's poor recognition capabilities concerning these signs are unacceptable and must be addressed to ensure safety.

\section{Reasoning}
\label{Reasoning}
Reasoning capabilities play a crucial role in autonomous driving systems, where not only basic decision-making is required but also the prediction of behaviors of other entities and the ability to respond promptly and accurately to emergencies and extreme scenarios, alongside route planning. This section details a series of evaluations conducted to assess GPT-4V's effectiveness in managing responses to emergency situations and extreme events, predicting the behaviors of other objects, and navigating dynamic environments.
\subsection{Emergency and extreme events}
\begin{figure*}[h]
\centering
\includegraphics[width=\textwidth]{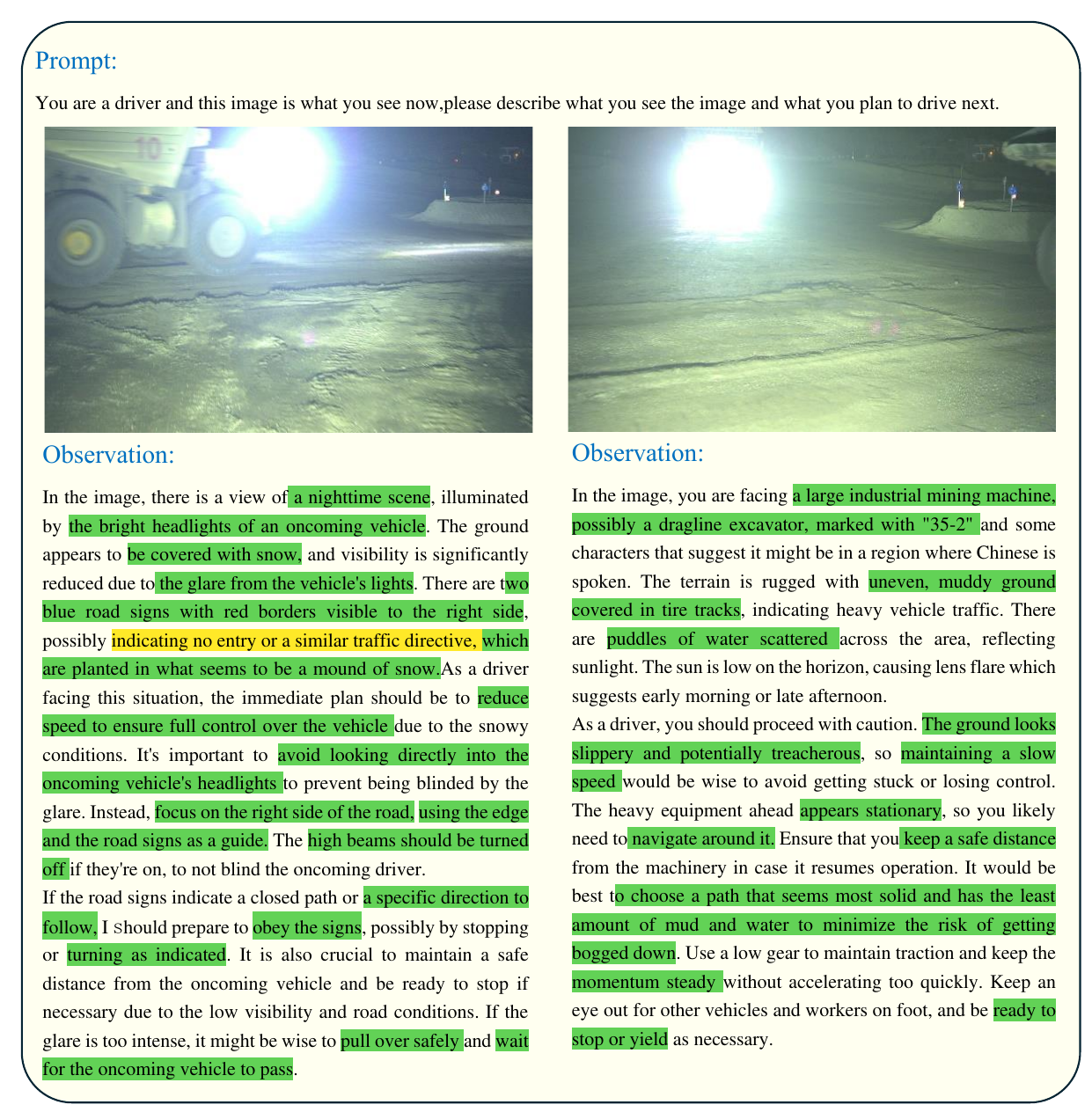}
\caption{\colorbox{green}{Green}highlights the right answer in understanding,\colorbox{red}{Red}highlights the wrong answer in understanding,\colorbox{yellow}{Yellow} highlights the incompetence in performing the task.}
\label{fig_36}
\end{figure*}

In scenarios with intense lighting, glare can significantly impair human vision, causing dazzlement and requiring time for drivers to adjust to changing light conditions. We selected two images to evaluate GPT-4V's performance in such extreme conditions in Figure~\ref{fig_36}. In the first scenario, where bright lighting intersected with vehicular traffic, GPT-4V successfully recognized a mining truck but erroneously identified the intense light as vehicle headlights, demonstrating its capability to recognize snow accumulation and road signs, even under night conditions. In this context, GPT-4V planned for the vehicle to cautiously follow the car ahead while maintaining a safe distance.In the second scenario, although GPT-4V continued to accurately identify roadside features and vehicle lights, it struggled with the accurate interpretation of road sign content. It executed evasive maneuvers to avoid an oncoming vehicle and deactivated the high beams, adapting its response based on the road sign indications. These tests highlighted GPT-4V's ability to navigate and plan routes effectively under challenging nighttime and intense light conditions.

\begin{figure*}[h]
\centering
\includegraphics[width=\textwidth]{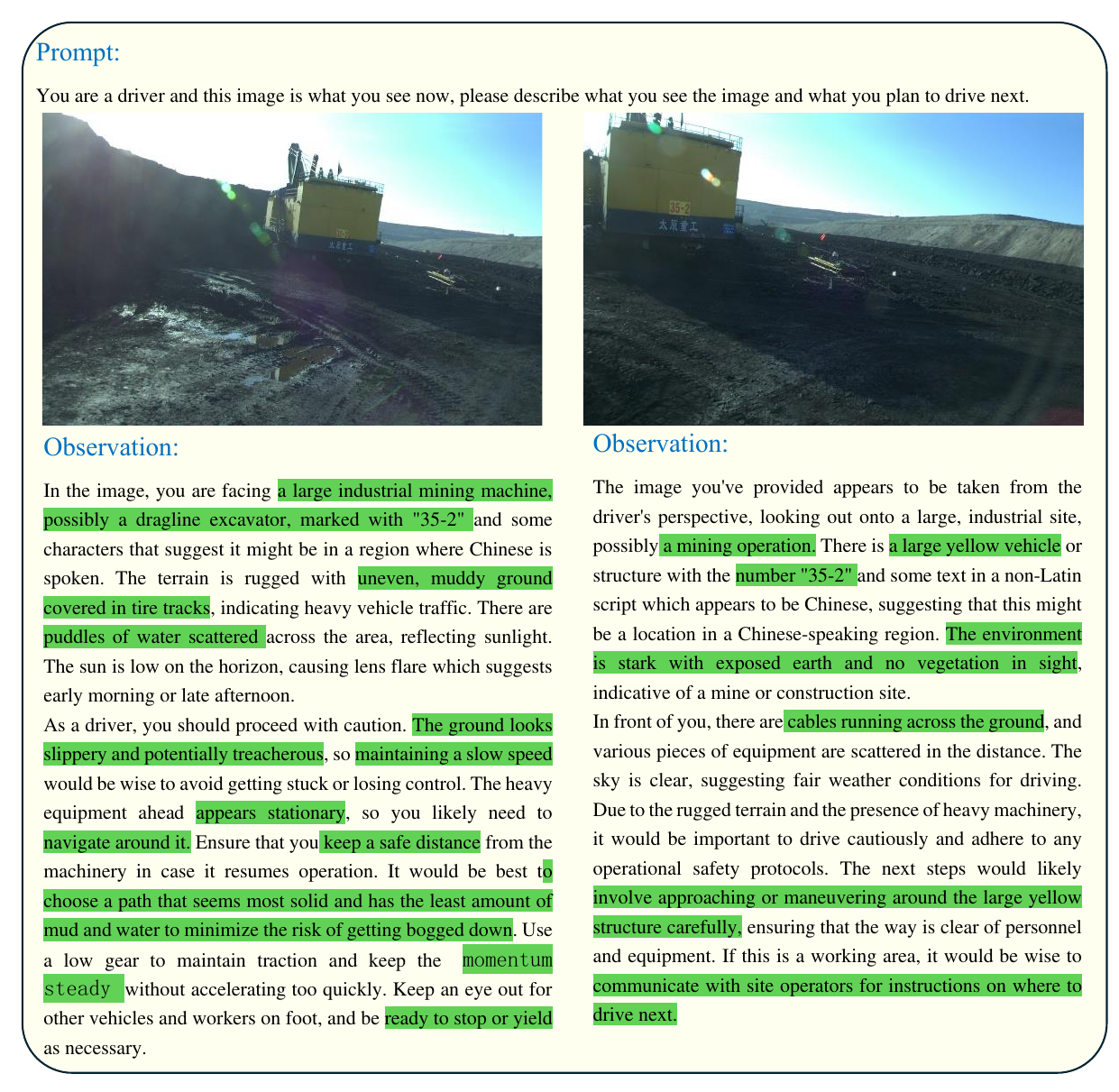}
\caption{\colorbox{green}{Green}highlights the right answer in understanding,\colorbox{red}{Red}highlights the wrong answer in understanding,\colorbox{yellow}{Yellow} highlights the incompetence in performing the task.}
\label{fig_37}
\end{figure*}

In challenging scenarios, the capability of GPT-4V to devise a suitable route that allows a vehicle to exit a specific area is under scrutiny.As depicted in Figure~\ref{fig_37},in the first image, GPT-4V's environmental assessment is largely accurate, showing a large excavator without a visible road. The system advises the vehicle to slow down and navigate around the machinery, but it fails to provide a clear detour path. In the second image, GPT-4V accurately evaluates the environment once more, instructing the vehicle to circumvent the large excavator ahead and to communicate with the onsite personnel. This tests illustrate that while GPT-4V can generally evaluate the environment, it struggles with detailed route planning.

\subsection{Temporal Sequences}

In this subsection, we evaluated GPT-4V's proficiency in understanding temporal sequences by presenting it with multiple series of images, each series labeled with sequence numbers or time stamps. This test aimed to examine GPT-4V's ability to interpret key frames effectively, focusing specifically on discerning the intentions behind other vehicles' movements. Additionally, we assessed the appropriateness of the subsequent driving actions taken by our vehicle based on GPT-4V's interpretations. This methodology was designed to determine GPT-4V’s capability to process and respond to dynamic situations in a continuous context, which is essential for making informed and predictive driving decisions.
In this sequence of images, the vehicle we monitored remained stationary. The sequence included specific scenarios such as construction sites with bulldozers present. GPT-4V successfully identified these elements and provided reasoned interpretations, correctly ascertaining the current driving behaviors executed by our vehicle. Generally, the assessments were accurate. However, some inaccuracies were noted—for instance, GPT-4V erroneously perceived that our vehicle was moving slowly, whereas in reality, it was the other vehicles that were in motion. This issue underscores the need for further refinement in GPT-4V's ability to accurately interpret dynamic scenes, particularly in distinguishing between the movements of various entities within the same visual context.

\begin{figure*}[h]
\centering
\includegraphics[width=\textwidth]{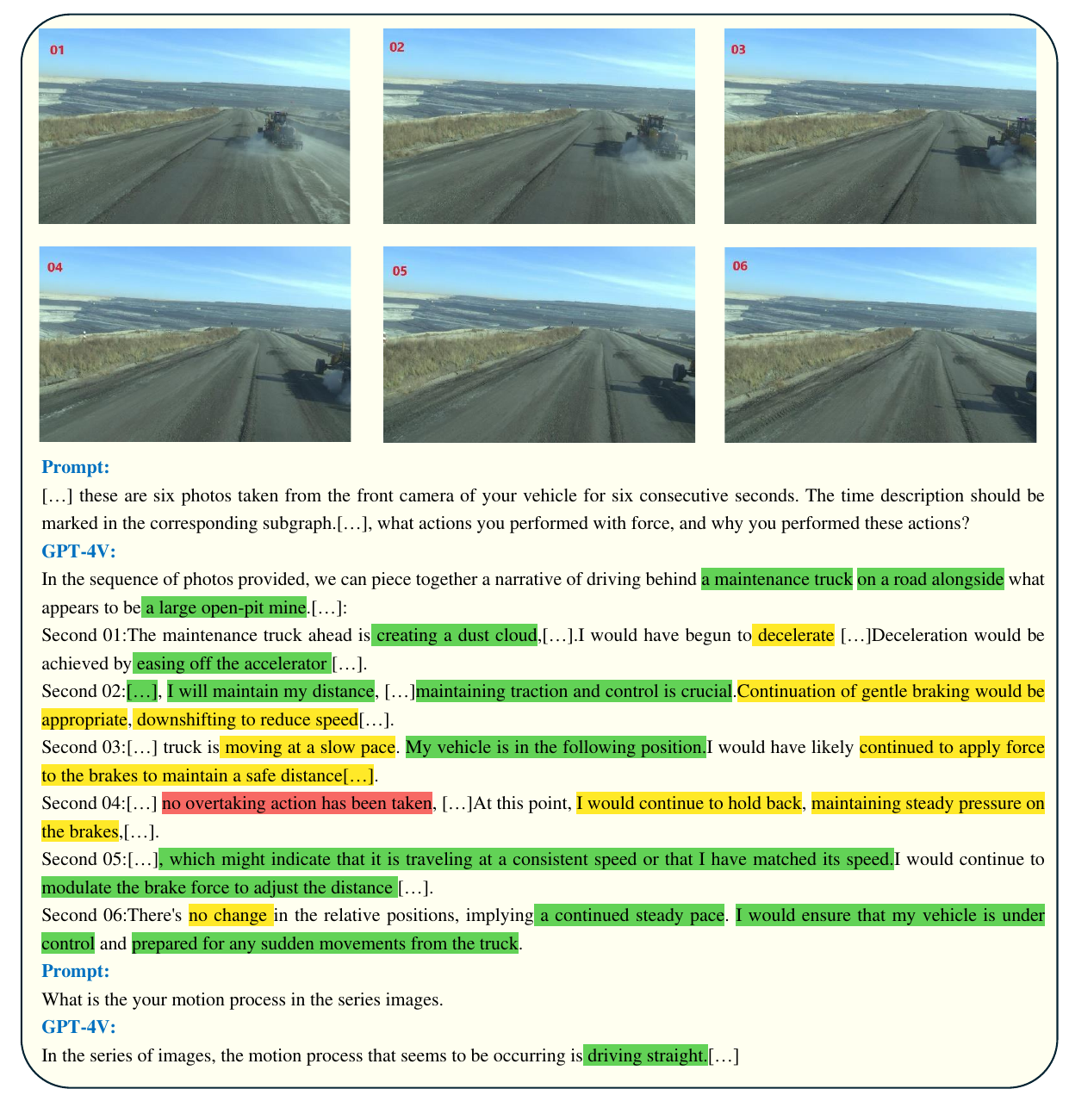}
\caption{\colorbox{green}{Green}highlights the right answer in understanding,\colorbox{red}{Red}highlights the wrong answer in understanding,\colorbox{yellow}{Yellow} highlights the incompetence in performing the task.}
\label{fig_38}
\end{figure*}

In Figure~\ref{fig_38}, GPT-4V is capable of deducing that this scenario involves an overtaking maneuver. While it does not precisely identify the type of the vehicle ahead, it accurately assesses the speed and direction of the preceding vehicle. However, its inference regarding the driving behavior of our vehicle is inappropriate; we anticipate that our vehicle should accelerate to overtake, yet GPT-4V advises deceleration. Moreover, GPT-4V erroneously assesses the state of our vehicle, which is actively engaged in overtaking.

\begin{figure*}[h]
\centering
\includegraphics[width=\textwidth]{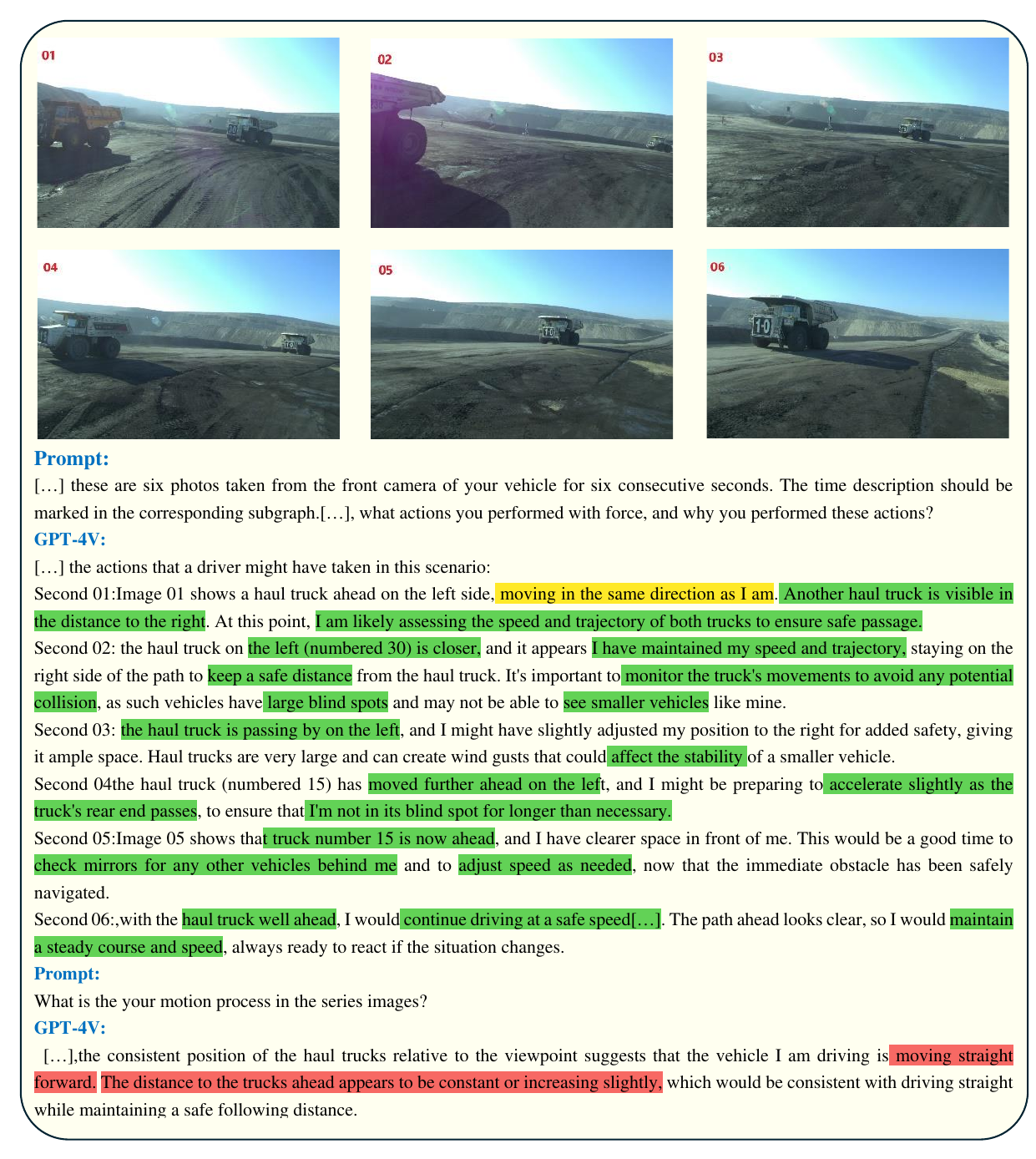}
\caption{\colorbox{green}{Green}highlights the right answer in understanding,\colorbox{red}{Red}highlights the wrong answer in understanding,\colorbox{yellow}{Yellow} highlights the incompetence in performing the task.}
\label{fig_39}
\end{figure*}

In Figure~\ref{fig_39}, which captures a turning maneuver involving several mining trucks, GPT-4V accurately assessed the status of most vehicles within the scene, though it did make a minor error in determining one vehicle's direction. However, it significantly misjudged the status of our vehicle. Errors included incorrect interpretations of our vehicle's turning action and its interactions with other vehicles. This indicates that GPT-4V's capability to deliver precise judgments in complex scenarios, particularly in understanding dynamic interactions between multiple vehicles, requires further enhancement.

\begin{figure*}[h]
\centering
\includegraphics[width=\textwidth]{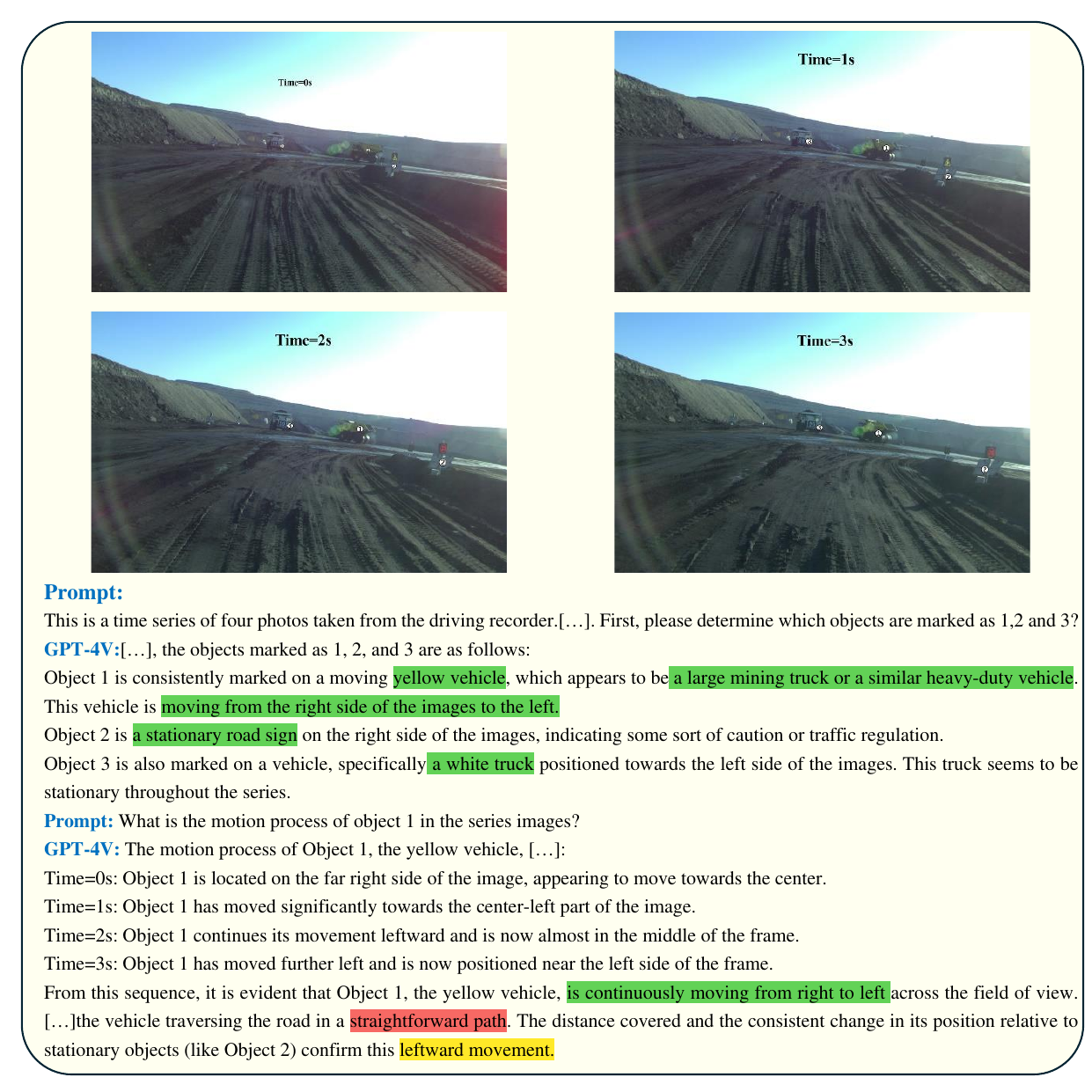}
\caption{\colorbox{green}{Green}highlights the right answer in understanding,\colorbox{red}{Red}highlights the wrong answer in understanding,\colorbox{yellow}{Yellow} highlights the incompetence in performing the task.}
\label{fig_40}
\end{figure*}

Figure~\ref{fig_40} displays a video captured at a road intersection featuring several trucks and traffic signals. Present are a white truck and a yellow truck near the intersection. GPT-4V successfully identifies the yellow truck, the white truck, and the road sign. However, while the yellow truck is positioned at a fork in the road and executing a turn, GPT-4V inaccurately interprets its movement. Despite this error with the yellow truck, GPT-4V's inference concerning the status of our vehicle is essentially correct, indicating that while its vehicle identification is reliable, its interpretation of dynamic vehicle movements at intersections needs improvement.

\begin{figure*}[h]
\centering
\includegraphics[width=\textwidth]{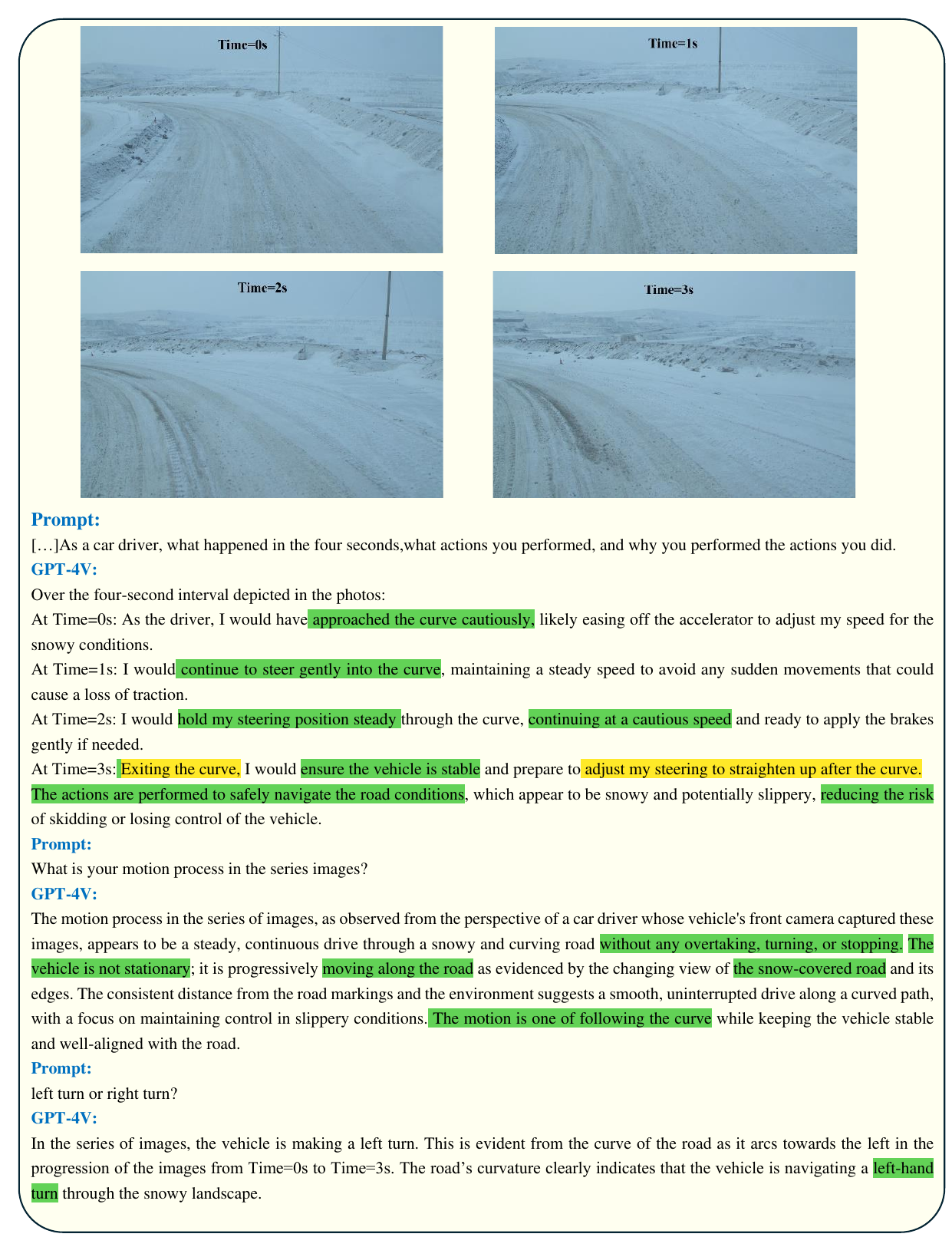}
\caption{\colorbox{green}{Green}highlights the right answer in understanding,\colorbox{red}{Red}highlights the wrong answer in understanding,\colorbox{yellow}{Yellow} highlights the incompetence in performing the task.}
\label{fig_41}
\end{figure*}

Figure~\ref{fig_41} depicts a vehicle navigating a turn on a snowy road. In such conditions, visibility is markedly reduced and the roadway becomes slick, necessitating all vehicles to reduce speed. The vehicle is shown moving slowly through the curve, and there are no visible obstructions like trucks or pedestrians. GPT-4V accurately identifies the snowy weather conditions and the curving road, correctly inferring that the vehicle needs to decelerate and initiate a left turn. However, GPT-4V incorrectly concludes that the vehicle has exited the curve. This error highlights a need for improvement in GPT-4V's ability to accurately track vehicle trajectories in challenging weather conditions.
\section{Act as A Driver}
\label{sec:driverAgent}
This study evaluates the decision-making capabilities of the GPT-4V GPT-4V to determine if they are on par with those of human drivers by conducting tests across five distinct driving tasks. These tasks—U-turns, overtaking, lane merging, obstacle avoidance, and parking—are selected to challenge the full range of the GPT-4V's autonomous driving capabilities within complex scenarios. These exercises necessitate that GPT-4V exhibit advanced skills in perception, planning, and decision-making. Vehicle speed and other relevant data were continuously provided to GPT-4V, which was responsible for generating appropriate observational and driving responses. The experimental results reveal that GPT-4V possesses robust decision-making abilities and can generally perform the functions of a driver effectively. However, it occasionally encounters issues, such as selecting incorrect routes due to the complexity of the scenario, highlighting areas for potential improvement in its operational algorithms.
\subsection{U-turning}
\begin{figure*}[h]
\centering
\includegraphics[width=\textwidth]{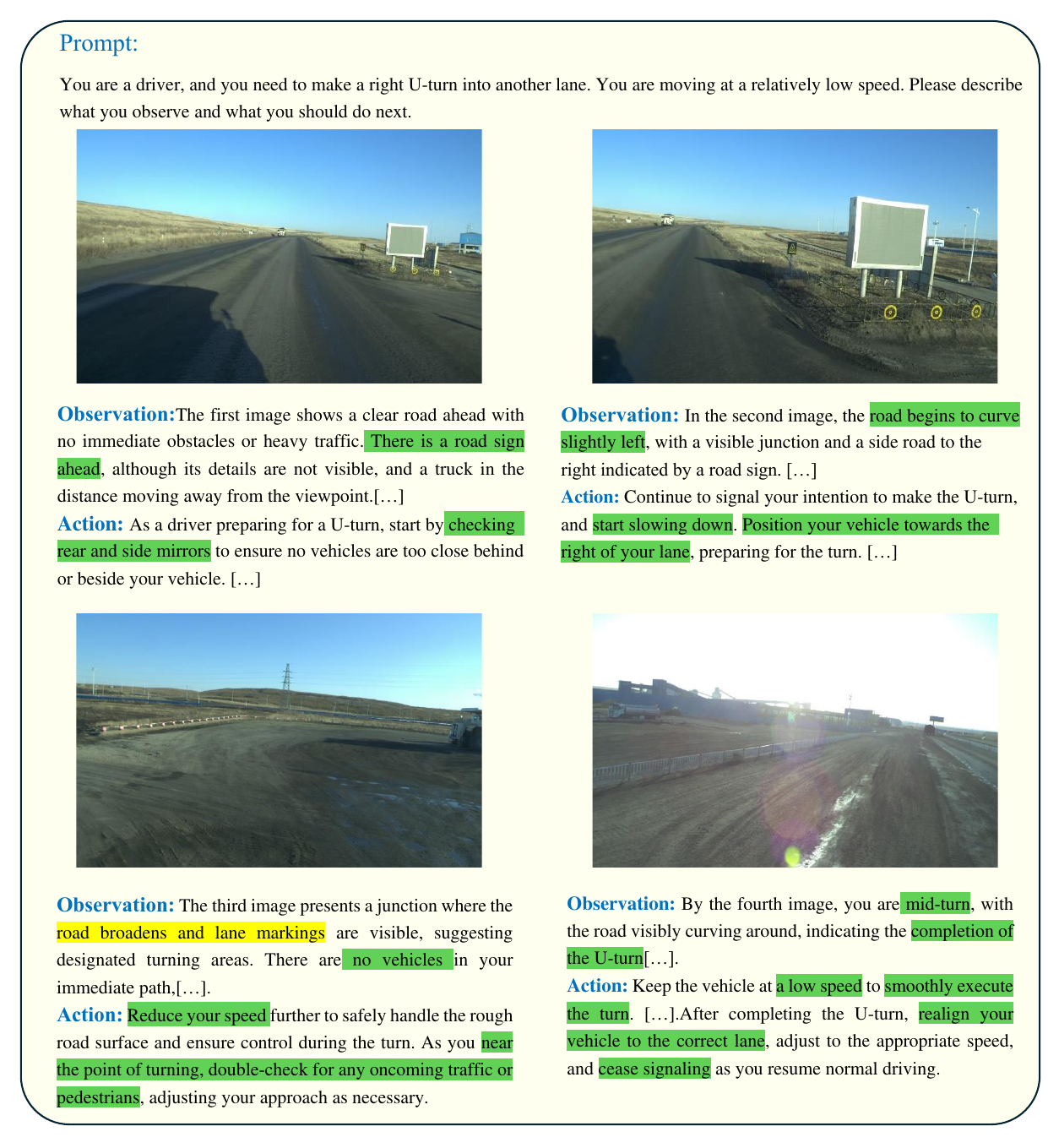}
\caption{\colorbox{green}{Green}highlights the right answer in understanding,\colorbox{red}{Red}highlights the wrong answer in understanding,\colorbox{yellow}{Yellow} highlights the incompetence in performing the task.}
\label{fig_42}
\end{figure*}

This study involves a scenario in which a vehicle performs a U-turning, a maneuver that requires careful consideration of both oncoming and following traffic, compounded by the low visibility conditions typical of mining areas. The ability of GPT-4V to manage this maneuver is critically assessed in a mining setting as illustrated in Figure~\ref{fig_42}.
In the initial frame, GPT-4V detects road signs and advises a speed reduction and a rear-view mirror check. In the subsequent frame, recognizing the rightward turn, it instructs the vehicle to slow down, prepare for the maneuver, and remain mindful of other road users. By the third frame, as the vehicle approaches the turning point, GPT-4V recommends proceeding with heightened caution. In the final frame, GPT-4V observes that the vehicle is nearing completion of the U-turning and directs a return to normal driving post-maneuver.
This evaluation demonstrates GPT-4V's sound understanding, reasoning, and decision-making throughout the U-turning process. It shows particular attentiveness to traffic dynamics and ensures the maneuver's safety by adhering to traffic regulations, affirming the GPT-4V's capability to navigate complex driving tasks effectively.

\subsection{Overtaking}
\begin{figure*}[h]
\centering
\includegraphics[width=\textwidth]{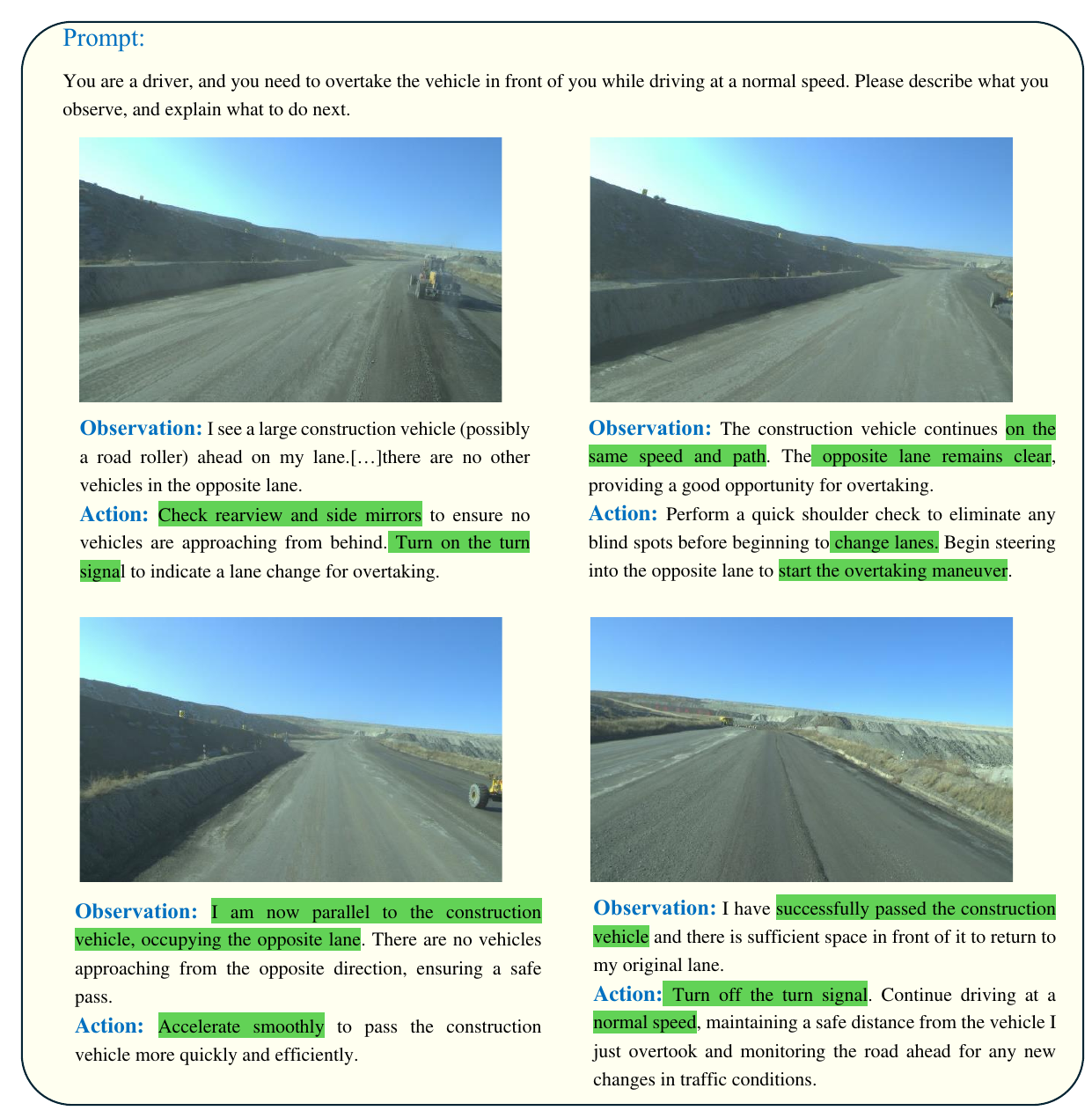}
\caption{\colorbox{green}{Green}highlights the right answer in understanding,\colorbox{red}{Red}highlights the wrong answer in understanding,\colorbox{yellow}{Yellow} highlights the incompetence in performing the task.}
\label{fig_43}
\end{figure*}

In this section, we evaluate the overtaking capabilities of GPT-4V, as depicted in Figure~\ref{fig_43}. The scenario involves our vehicle driving on an unpaved road with a construction vehicle ahead that we intend to overtake.

In the initial frame, GPT-4V identifies the construction vehicle and assesses the safety conditions for initiating an overtaking maneuver. By the second frame, observing that the construction vehicle maintains a consistent speed and direction, GPT-4V commands a lane change to begin the overtaking process. In the third frame, as our vehicle runs parallel to the construction vehicle, GPT-4V instructs an acceleration to complete the overtaking. In the final frame, GPT-4V confirms that our vehicle has successfully overtaken the construction vehicle and has returned to its lane, advising a continuation of normal driving.

This example demonstrates GPT-4V’s strong overtaking capabilities, showcasing precise observation and sound decision-making. Remarkably, GPT-4V can effectively differentiate between lanes on unstructured roads that lack clear lane markings, demonstrating a level of performance that exceeds most semantic segmentation algorithms.

\subsection{Pathfinding}
\begin{figure*}[h]
\centering
\includegraphics[width=\textwidth]{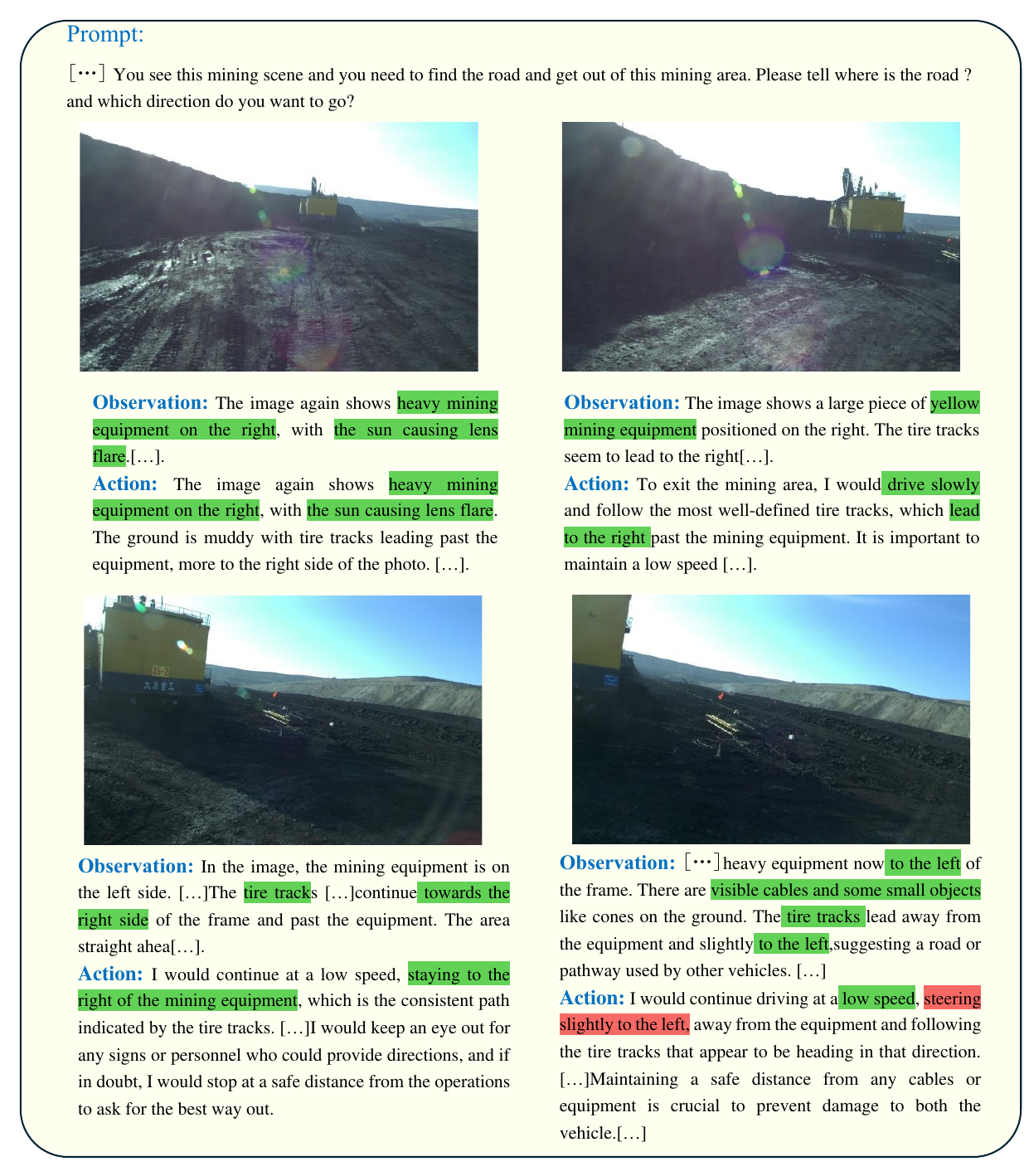}
\caption{\colorbox{green}{Green}highlights the right answer in understanding,\colorbox{red}{Red}highlights the wrong answer in understanding,\colorbox{yellow}{Yellow} highlights the incompetence in performing the task.}
\label{fig_44}
\end{figure*}

In this section, we assess the pathfinding capabilities of GPT-4V within the complex environment of a mining site, characterized by heavy machinery, various construction equipment, and a lack of clear lanes or signage for navigation, as depicted in Figure~\ref{fig_44}.

In the first frame, GPT-4V identifies a yellow excavator positioned to the right of the scene and instructs the vehicle to slow down, follow the tire tracks, and move towards the right of the excavator. In the second frame, GPT-4V notes the muddy road and rough terrain, suggesting that the exit might be to the right. In the third frame, upon detecting sunspots, GPT-4V advises the vehicle to make a right turn while maintaining a low speed to navigate safely. However, in the fourth frame, GPT-4V observes scattered small objects and tire tracks leading left and mistakenly concludes that the vehicle should turn left.
Despite not accurately identifying the correct path leading out of the mining site, GPT-4V's performance was notable. It successfully discerned the general direction and recommended appropriate speeds for navigating through an area devoid of clear paths or signs, demonstrating its potential in complex environment navigation.

\subsection{Parking}
\begin{figure*}[h]
\centering
\includegraphics[width=\textwidth]{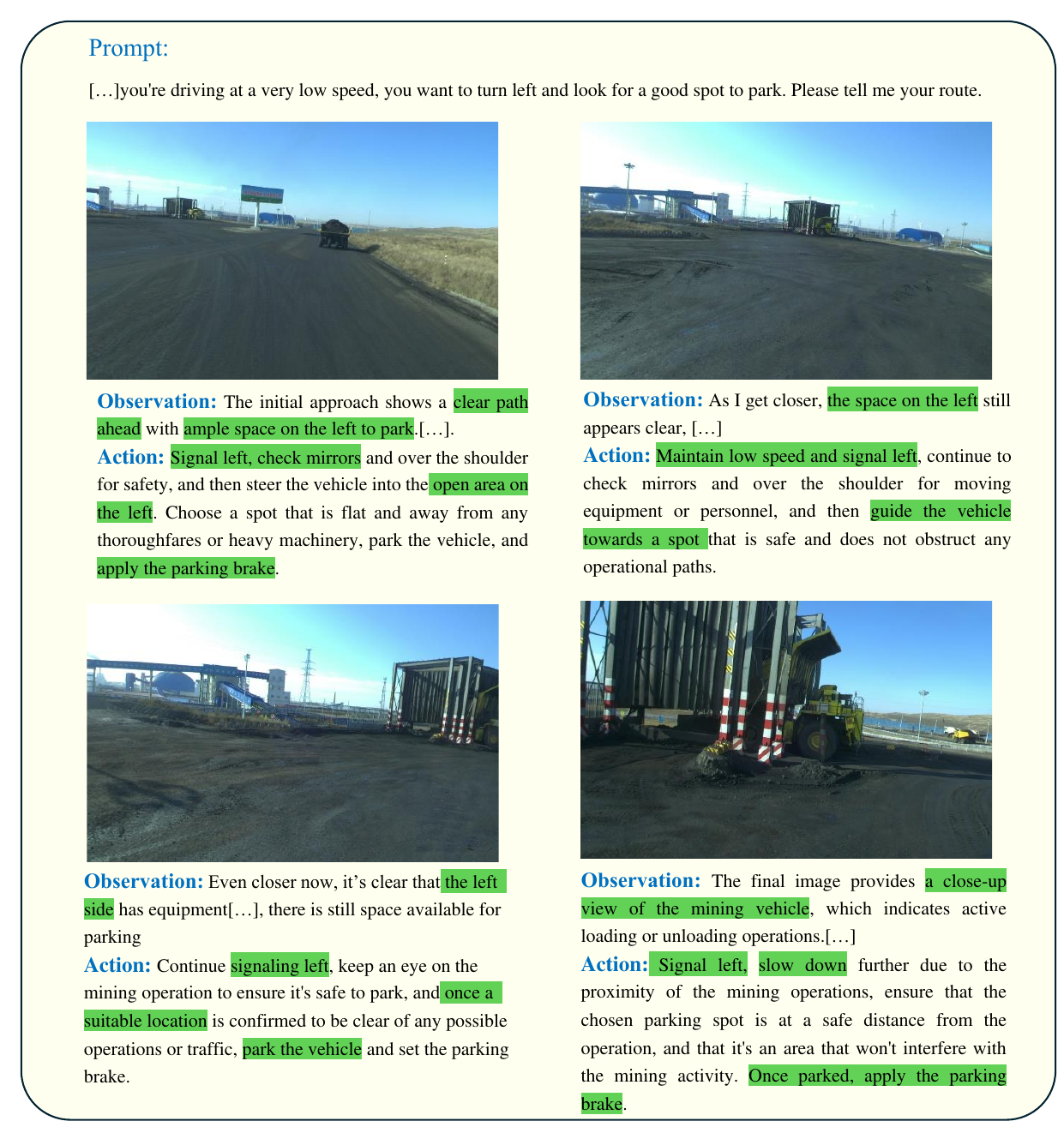}
\caption{\colorbox{green}{Green}highlights the right answer in understanding,\colorbox{red}{Red}highlights the wrong answer in understanding,\colorbox{yellow}{Yellow} highlights the incompetence in performing the task.}

\label{fig_45}
\end{figure*}

In this section, we assess the parking capabilities of the vehicle within a mining area where no clearly defined parking zones exist, necessitating vehicles to independently locate suitable parking spaces.
As illustrated in Figure~\ref{fig_45}, in the initial frame, GPT-4V identifies ample space to the left suitable for parking and instructs the vehicle to turn left, activating the left turn signal to initiate the parking maneuver. In the second frame, GPT-4V continues to guide the vehicle left towards the identified parking spot. By the third frame, GPT-4V directs further leftward movement, preparing the vehicle for the parking process. In the final frame, noting a large mining truck in proximity, GPT-4V advises slowing down while completing the left turn and then stopping to park.
This scenario demonstrates GPT-4V’s effectiveness in locating and navigating to an appropriate open area for parking within the complex environment of a mining site, executing the task with precision and without significant errors.

\subsection{Lane Merging}
\begin{figure*}[h]
\centering
\includegraphics[width=\textwidth]{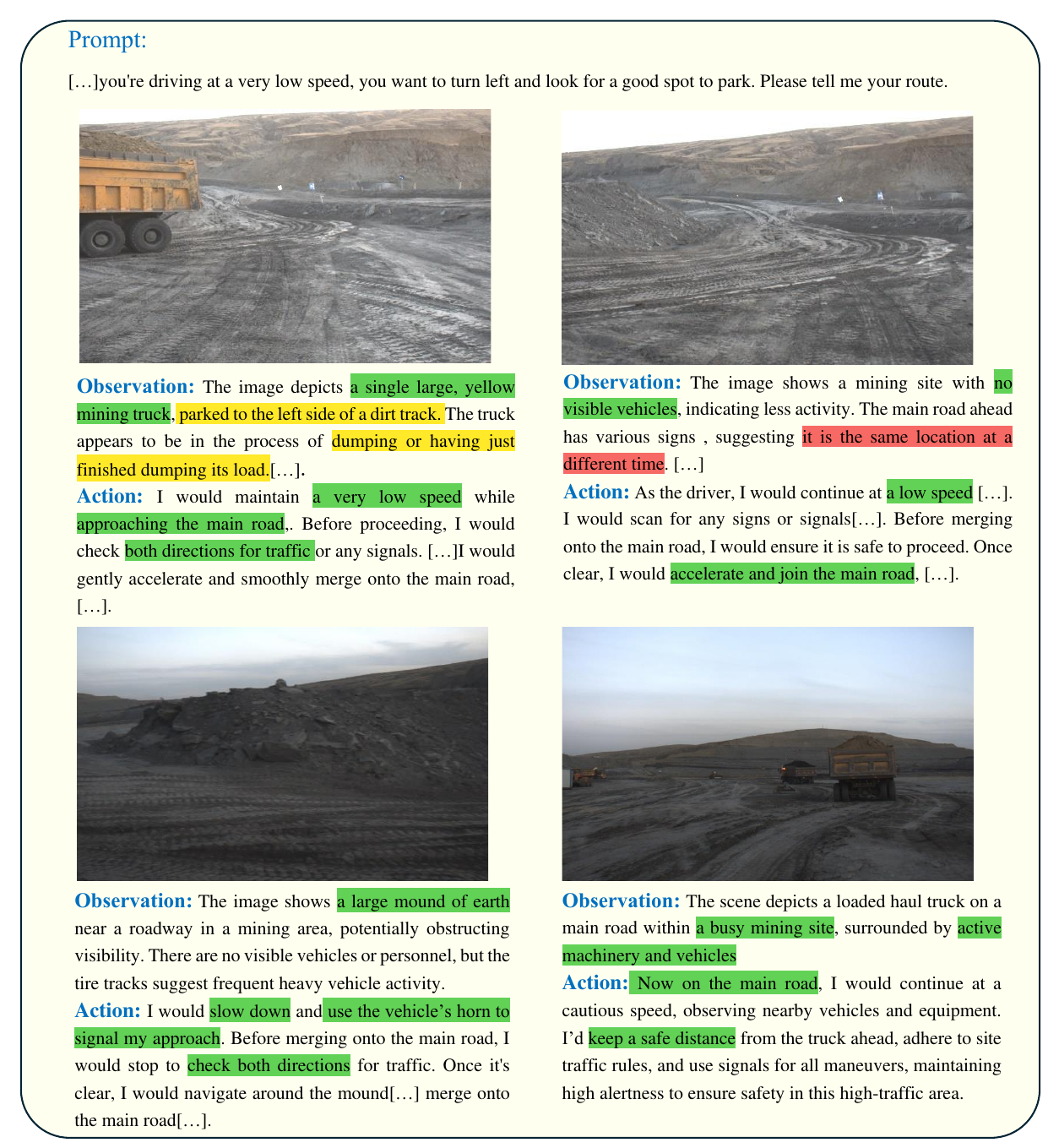}
\caption{\colorbox{green}{Green}highlights the right answer in understanding,\colorbox{red}{Red}highlights the wrong answer in understanding,\colorbox{yellow}{Yellow} highlights the incompetence in performing the task.}
\label{fig_46}
\end{figure*}

In this section, we evaluated GPT-4V capability to guide vehicles in merging onto main roads in Figure~\ref{fig_46}, focusing on a scenario at an intersection leading to a main road.
In the first frame, observing a mining truck, muddy roads, and tire tracks, GPT-4V  directed the vehicle to decelerate, proceed with caution, keep an eye on the vehicle ahead, and remain aware of the surrounding environment. In the second frame, GPT-4V  inaccurately assessed the vehicle's status, mistakenly believing it was stationary when it was actually moving. It then instructed the vehicle to accelerate to facilitate merging onto the main road. In the third frame, the presence of a large mound obstructing the view led GPT-4V  to activate the vehicle's horn to signal its approach and to help navigate around the obstruction. By the fourth frame, GPT-4V determined that the vehicle had successfully reached the main road and recommended continuing to drive cautiously.
This scenario demonstrates that our vehicle was directed to wait for other vehicles to pass before cautiously merging onto the main road. GPT-4V environmental assessments and commands regarding the vehicle's maneuvering were largely accurate, despite a notable error in perceiving the vehicle's motion status, which was indeed in motion throughout the sequence.

\section{Conclusion}
This study provides a comprehensive evaluation of the GPT-4V model in the context of autonomous driving within mining environments. It focuses on the model’s abilities in scene understanding, reasoning, and performing driving tasks typical of human drivers. GPT-4V excelled in recognizing and interpreting a range of environmental elements, including vehicles, pedestrians, and road signs. Despite these strengths, it encountered difficulties in accurately identifying specific vehicle types and handling dynamic interactions.

In scenarios featuring emergency and extreme conditions, GPT-4V showcased substantial reasoning capabilities, effectively navigating through challenges presented by intense lighting and limited visibility. Nonetheless, the model occasionally made errors in interpreting vehicle movements and road sign information. When tasked with practical driving operations such as U-turnings, overtaking, lane merging, pathfinding, and parking, GPT-4V generally performed well in maneuvering and decision-making. However, it faced challenges in complex scenarios, particularly in accurately tracking vehicle trajectories and selecting the best routes.

The assessment pointed out key areas needing improvement. There is a critical need for GPT-4V to enhance its accuracy in vehicle recognition and counting, especially in conditions of poor visibility. Moreover, the model must improve its ability to interpret and predict the dynamic interactions and movements of multiple vehicles. While it was competent in identifying stationary objects and static scenes, its understanding of dynamic scenes, such as those involving moving mechanical arms or changing road conditions, proved less dependable.

Concluding, GPT-4V holds significant promise in advancing autonomous driving technology within mining environments, attributed to its capacity to manage complex scenes and implement strategic driving maneuvers. Nevertheless, achieving reliable and safe operations necessitates addressing the identified deficiencies in vehicle recognition, interaction interpretation, and dynamic scene comprehension.

\clearpage
\section{Appendix}

\begin{figure*}[h]
\centering
\includegraphics[width=\linewidth]{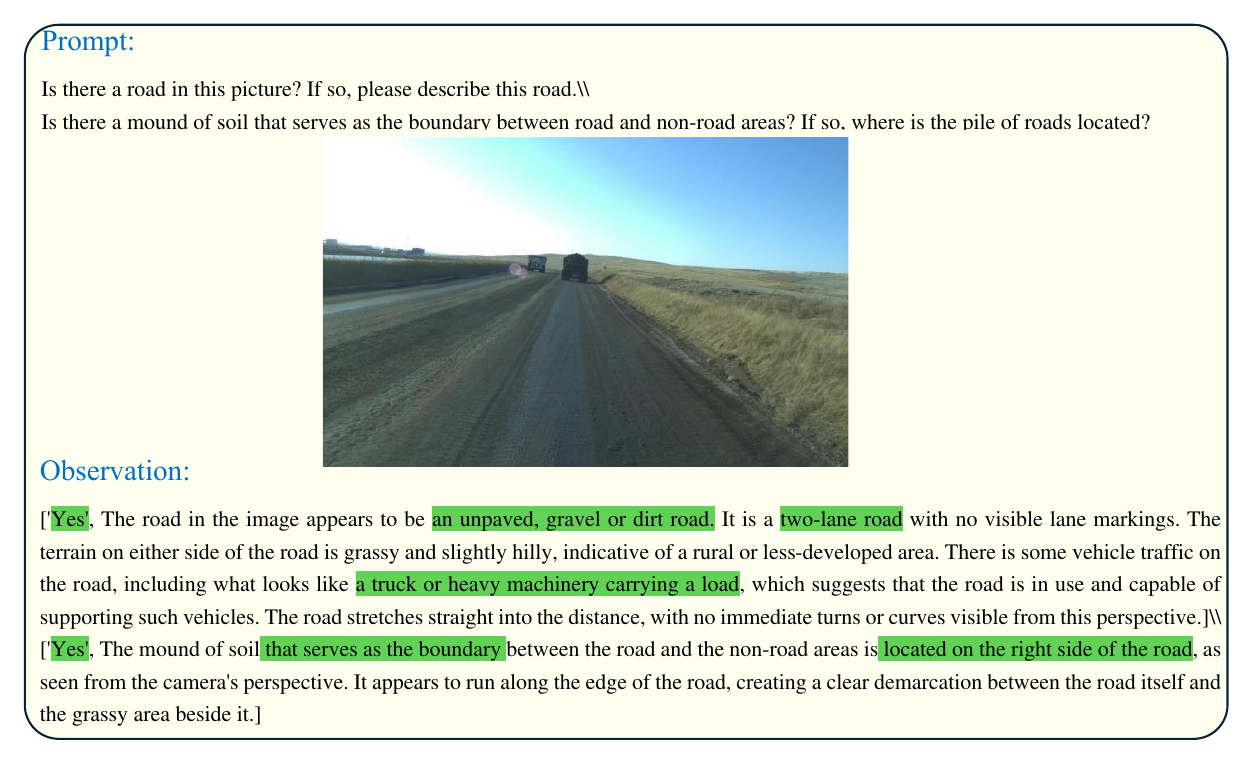}
\caption{\colorbox{green}{Green}highlights the right answer in understanding,\colorbox{red}{Red}highlights the wrong answer in understanding,\colorbox{yellow}{Yellow} highlights the incompetence in performing the task.}
\label{fig_15}
\end{figure*}

\begin{figure*}[h]
\centering
\includegraphics[width=\linewidth]{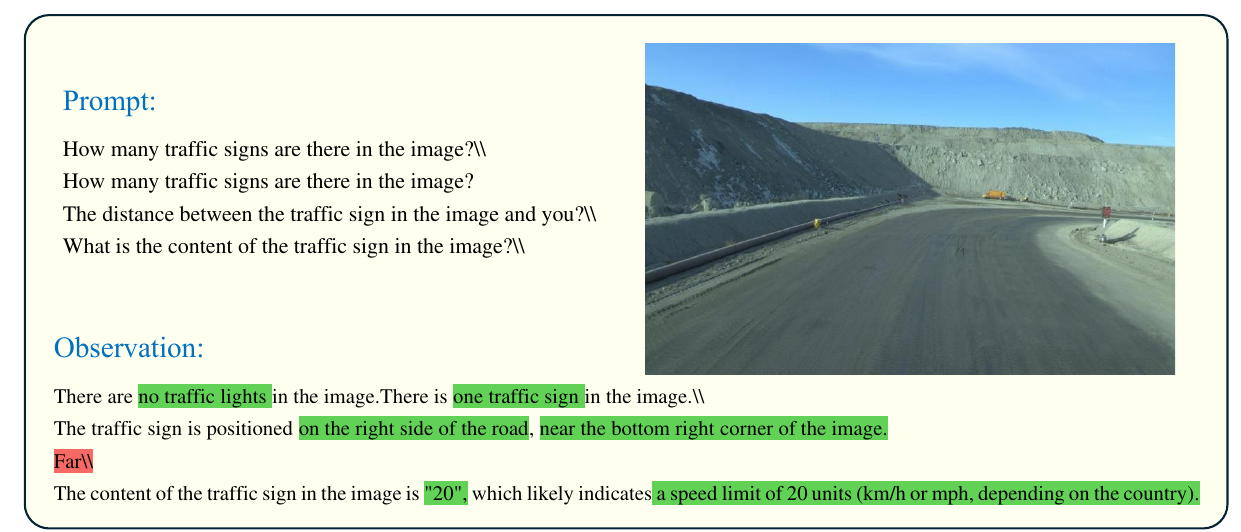}
\caption{\colorbox{green}{Green}highlights the right answer in understanding,\colorbox{red}{Red}highlights the wrong answer in understanding,\colorbox{yellow}{Yellow} highlights the incompetence in performing the task.}
\label{fig_23}
\end{figure*}

\begin{figure*}[h]
\centering
\includegraphics[width=\linewidth]{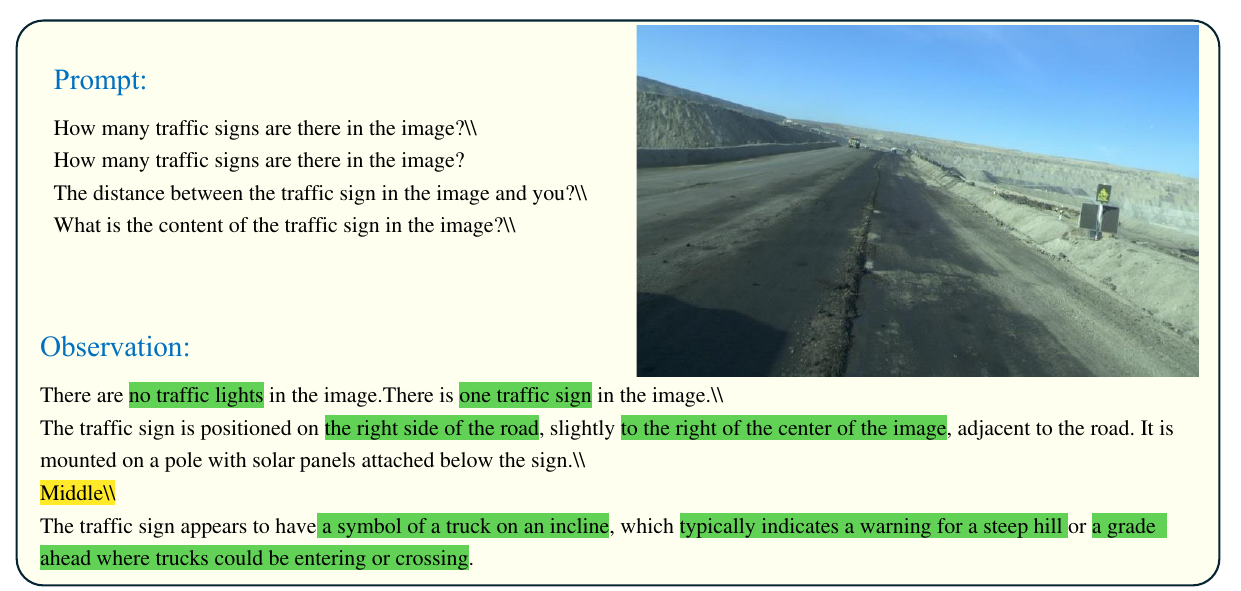}
\caption{\colorbox{green}{Green}highlights the right answer in understanding,\colorbox{red}{Red}highlights the wrong answer in understanding,\colorbox{yellow}{Yellow} highlights the incompetence in performing the task.}
\label{fig_24}
\end{figure*}

\begin{figure*}[h]
\centering
\includegraphics[width=\linewidth]{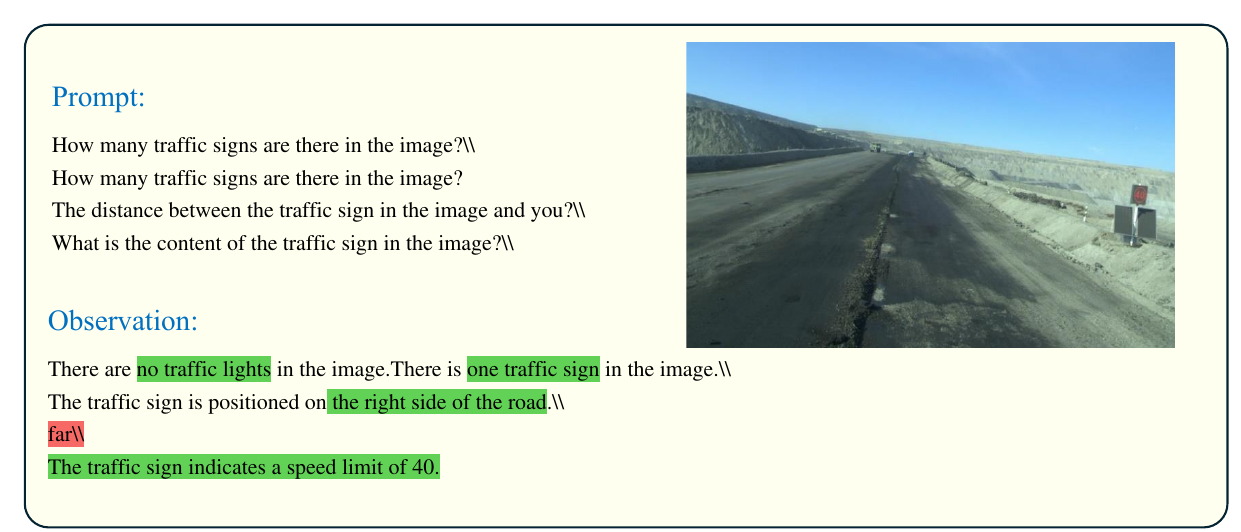}
\caption{\colorbox{green}{Green}highlights the right answer in understanding,\colorbox{red}{Red}highlights the wrong answer in understanding,\colorbox{yellow}{Yellow} highlights the incompetence in performing the task.}
\label{fig_25}
\end{figure*}

\begin{figure*}[h]
\centering
\includegraphics[width=\linewidth]{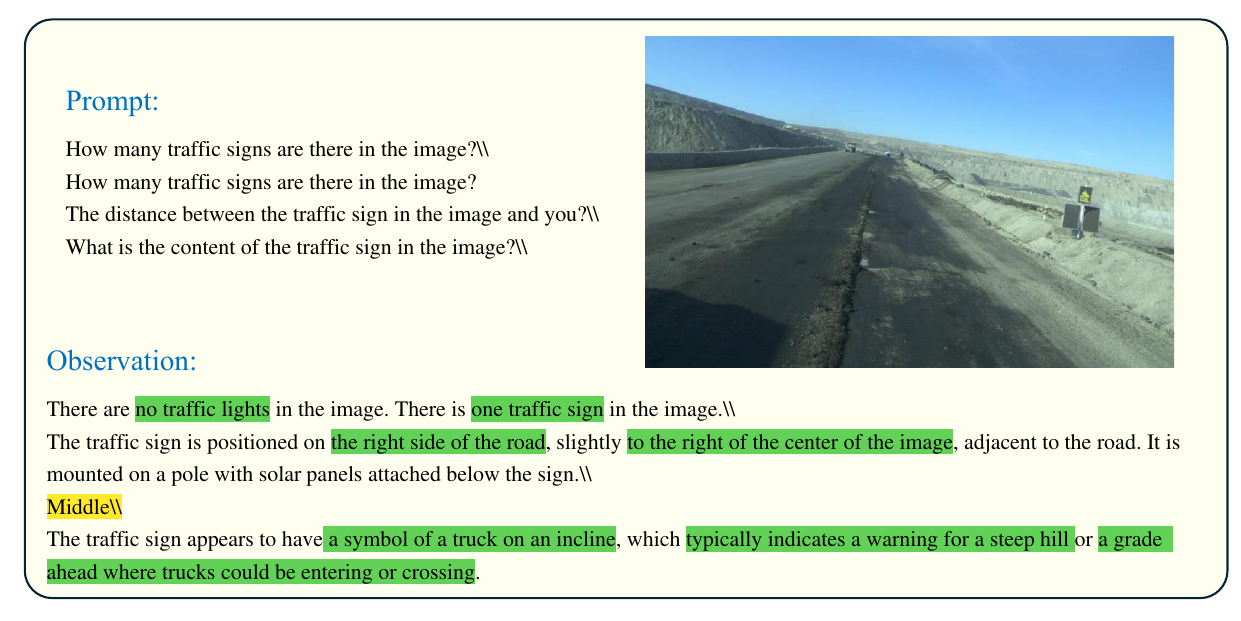}
\caption{\colorbox{green}{Green}highlights the right answer in understanding,\colorbox{red}{Red}highlights the wrong answer in understanding,\colorbox{yellow}{Yellow} highlights the incompetence in performing the task.}
\label{fig_26}
\end{figure*}

\begin{figure*}[h]
\centering
\includegraphics[width=\linewidth]{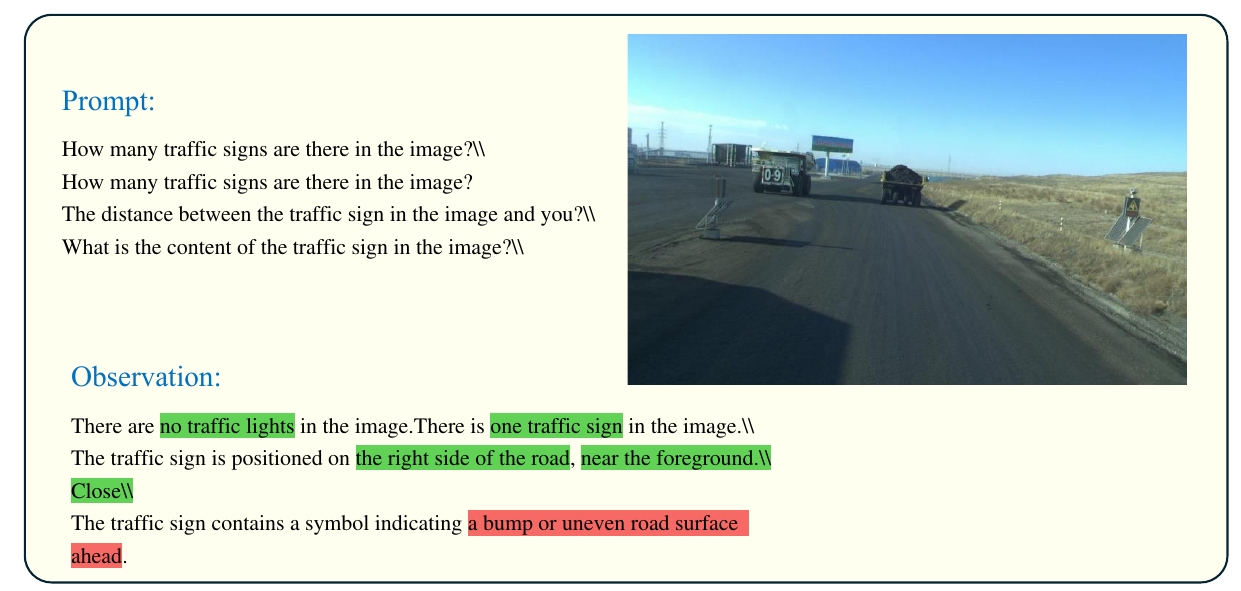}
\caption{\colorbox{green}{Green}highlights the right answer in understanding,\colorbox{red}{Red}highlights the wrong answer in understanding,\colorbox{yellow}{Yellow} highlights the incompetence in performing the task.}
\label{fig_28}
\end{figure*}

\begin{figure*}[h]
\centering
\includegraphics[width=\linewidth]{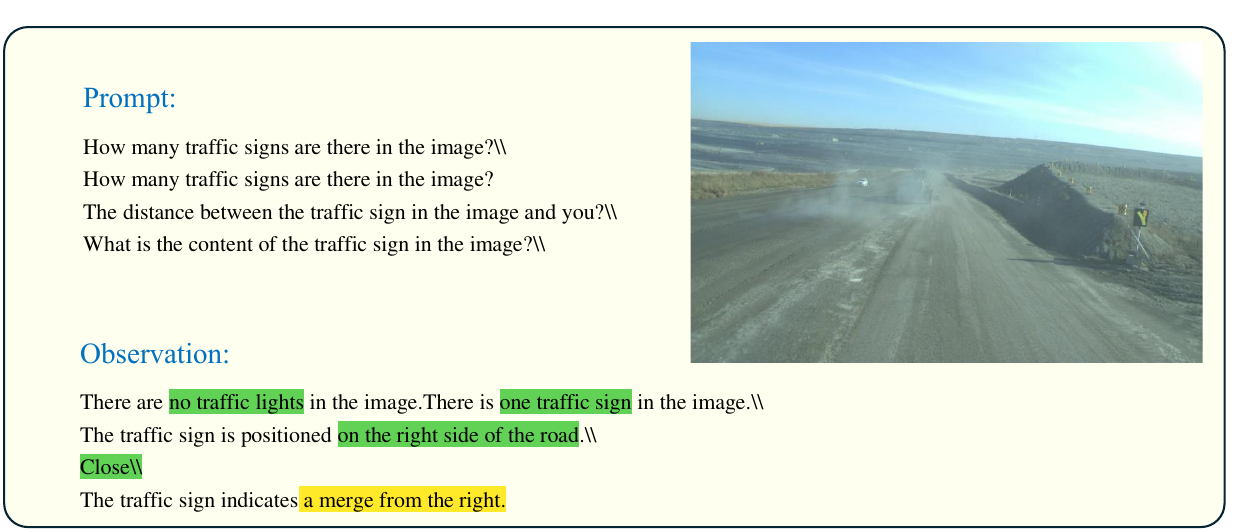}
\caption{\colorbox{green}{Green}highlights the right answer in understanding,\colorbox{red}{Red}highlights the wrong answer in understanding,\colorbox{yellow}{Yellow} highlights the incompetence in performing the task.}
\label{fig_31}
\end{figure*}

\clearpage
{
\bibliographystyle{plain}
\bibliography{egbib}
}

\end{document}